\documentclass[10pt,twocolumn,letterpaper]{article}

\usepackage[pagenumbers]{wacv} 

\usepackage{subcaption}
\usepackage{algorithm}
\usepackage{algpseudocode}
\usepackage{multirow}
\usepackage{enumitem}

\usepackage[accsupp]{axessibility}  

\definecolor{wacvblue}{rgb}{0.21,0.49,0.74}
\usepackage[pagebackref,breaklinks,colorlinks,allcolors=wacvblue]{hyperref}

\def\wacvPaperID{3454} 
\def\confName{WACV}
\def\confYear{2027}

\title{Loomis Painter: Learning Painting Processes by Reversing Image-to-Video Generation\vspace{-0.2cm}}

\author{
    Markus Pobitzer$^{1}$ \quad
    Chang Liu$^{1,}$\textsuperscript{\textdagger} \quad
    Chenyi Zhuang$^{1}$ \quad
    Teng Long$^{1}$ \quad
    Bin Ren$^{1,2}$ \quad
    Nicu Sebe$^{1}$ \\
    [0.3cm]
    $^{1}$University of Trento \qquad $^{2}$University of Pisa \\
    {\tt\small markus.pobitzer@unitn.it, \textsuperscript{\textdagger}Corresponding author} \\
}

\begin{document}

\twocolumn[{%
  \renewcommand\twocolumn[1][]{#1}%
  \maketitle
  \begin{center}
    \vspace{-7mm}
    \captionsetup{type=figure}

    \begin{subfigure}{\linewidth}
      \centering
      \includegraphics[width=\linewidth]{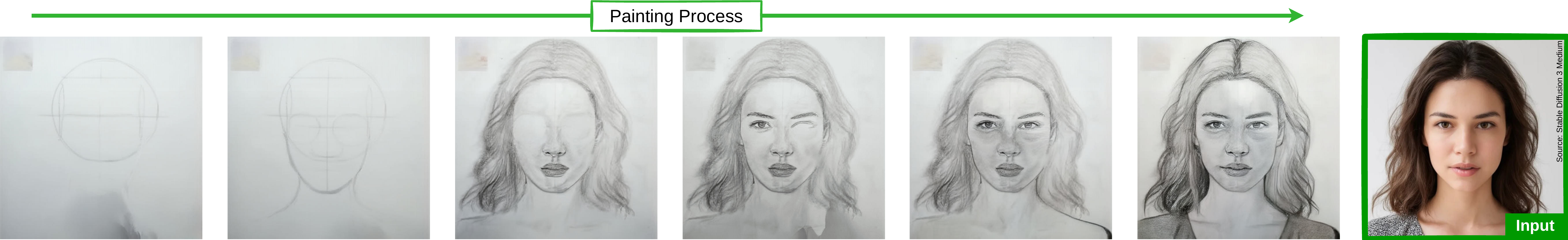}
      \caption{Loomis-style pencil portraits painting from any portrait photo.}
      \label{fig:teaser_a}
    \end{subfigure}
    \vspace{-3mm} 

    \begin{subfigure}{\linewidth}
      \centering
      \includegraphics[width=\linewidth]{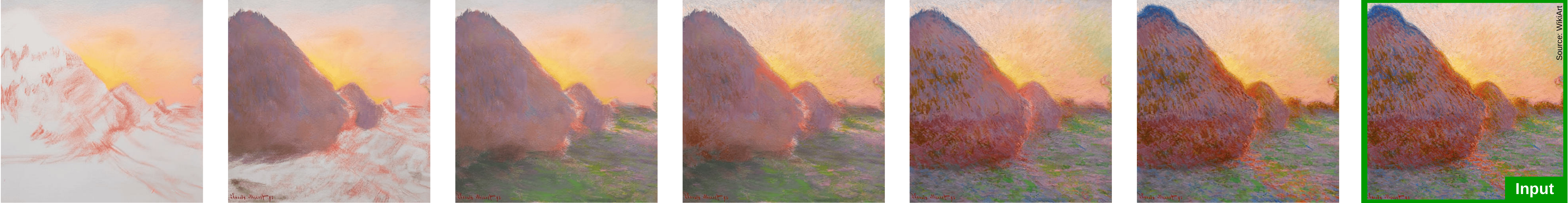}
      \caption{Reconstructing the painting process while preserving the input image.}
      \label{fig:teaser_b}
    \end{subfigure}
    \vspace{-3mm}

    \begin{subfigure}{\linewidth}
      \centering
      \includegraphics[width=\linewidth]{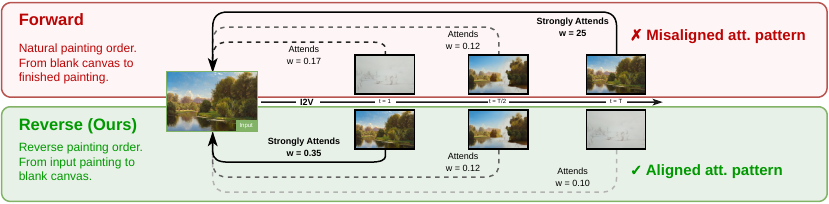}
      \caption{Forward painting misaligns I2V attention. Our solution: reverse painting.}
      \label{fig:teaser_c}
    \end{subfigure}

    \vspace{-2mm}

    \addtocounter{figure}{-1}
    \captionof{figure}{Loomis Painter: Our method reconstructs the painting process of artworks with supported media faithfully (b), or input images in a supported art media as in (a). The title is inspired by the Loomis portrait method, which we also enable. Our main finding: reverse painting order aligns better with image to video models (c). Images with green borders are inputs; all others are generated by our method.}
    \label{fig:teaser}
    \vspace{-1mm}
  \end{center}
}]


\begin{abstract}
Step-by-step painting tutorials are used for learning artistic techniques, but existing video resources lack interactivity and personalization. While recent generative models have advanced artistic image synthesis, they still struggle to generate temporally and structurally consistent videos.
Video models excel in consistent frame generation, yet fine-tuning an image-to-video model on forward painting sequences creates a long-range dependency on the first-frame reference, disrupting its pretrained attention pattern and hindering reconstruction. We address this issue by training on reversed sequences that better preserve the reference-attention behavior.
Building on this, we present Loomis Painter, a framework for reconstructing coherent painting processes.
We further introduce a media-transfer training strategy that decouples input appearance from the target process, enabling generation across different artistic media.
Trained on a curated dataset of real-world painting processes, our method outperforms existing baselines across LPIPS, DINO, and CLIP metrics and is consistently preferred in expert and non-expert user studies. Code available \href{https://github.com/Markus-Pobitzer/wlp}{here}.
\end{abstract}

\section{Introduction}
\label{sec:intro}

Sketching and painting are creative processes that are hard to master. Many tutorials exist as step-by-step guides in the form of books and videos. However, books cannot show the complete dynamic process, and existing instructional videos are inherently passive, lacking interactivity and personalized guidance. Replicating a scene also includes media-specific steps such as layering color with oil \cite{speed1987oil} and faithfully reconstructing proportions. A prominent example is the Loomis Method \cite{loomis2021drawing}, developed by Andrew Loomis, which demonstrates a structural approach to drawing a head with correct proportions.





\noindent
While recent generative approaches have made progress in reconstructing painting processes \cite{paintsundo, song2024processpainter, chen2024inverse}, generating temporally coherent sequences that faithfully recover a reference artwork remains challenging. Pretrained image-to-video (I2V) models provide a natural foundation for this task because of their strong temporal modeling capabilities. However, directly fine-tuning them on forward painting sequences introduces a fundamental mismatch with their pretrained behavior. The reference artwork is provided in the first frame, while its complete reconstruction is only reached at the end of the sequence. As a result, the model must maintain long-range dependence on the reference image throughout generation, substantially altering its pretrained reference-attention pattern.

\noindent We address this issue by training on reversed sequences, starting from the completed artwork and progressively removing content. This formulation places the reference artwork at the beginning of the generation trajectory and preserves the reference-attention behavior of the pretrained image-to-video model. After generation, the sequence is reversed to recover the forward painting process.

\noindent
Beyond temporal coherence, two additional challenges remain. First, existing works use narrowly scoped datasets: Inverse Painting \cite{chen2024inverse} uses limited data, ProcessPainter \cite{song2024processpainter} leverages synthetic samples, and PaintsUndo \cite{paintsundo} focuses on digital art. Curating datasets of traditional painting processes is challenging, as real-world videos suffer from hand occlusions, camera movements, and irrelevant footage.

\noindent
Second, existing methods do not support art media transfer. They cannot translate an input image into the painting process of a different medium, such as converting a real portrait photograph into a step-by-step Loomis pencil sketch.

\noindent In this work, we propose a unified framework for multi-media painting process generation built upon a video generation model. To capture authentic workflows, we construct a diverse dataset of traditional drawing and painting videos. To achieve this, we introduce an automated curation pipeline that isolates the painting process and crops the canvas while removing hand occlusions and visual clutter, enabling the model to learn accurate, unobstructed processes.

\noindent Next, we show that a reverse-painting training strategy aligns with the underlying video generation model. This formulation grounds the first generated frame directly on the input image, teaching the model to sequentially "unpaint" the canvas. 

\noindent Finally, we introduce a control mechanism that embeds multiple artistic media into the diffusion model's conditional space. This enables the model to capture medium-specific textures and stroke patterns within a shared latent space. As a result, our framework supports the temporally coherent translation of painting processes across different styles, what we term art media transfer.

\noindent We conduct experiments evaluating faithful reconstruction, temporal coherence, and final-image fidelity, achieving state-of-the-art performance. Furthermore, both a expert evaluation and a general human preference study demonstrate a clear preference for our approach over existing baselines. In summary, our contributions are:

\begin{itemize}[noitemsep,topsep=4pt,parsep=0pt,partopsep=0pt]
    \item \textbf{Loomis Painter:} We propose a image-to-video generation model specifically designed for step-by-step reconstruction of the painting process.

    \item \textbf{Reverse-Painting Strategy:} We identify a mismatch between forward painting generation and the pretrained reference-attention behavior of I2V models, and introduce a reverse-painting training strategy that better preserves this behavior, improving reconstruction fidelity.


    \item \textbf{Art Media Transfer:} We introduce a training strategy enabling generation across different artistic media, including Loomis-style pencil drawings from photographs.

    \item \textbf{Dataset Pipeline \& Evaluation:} We develop an automated pipeline for curating real-world painting-process videos and demonstrate state-of-the-art performance.
\end{itemize}

\section{Related Work}
\label{sec:related_work}

\begin{figure*}[ht]
  \centering
  \includegraphics[width=\linewidth]{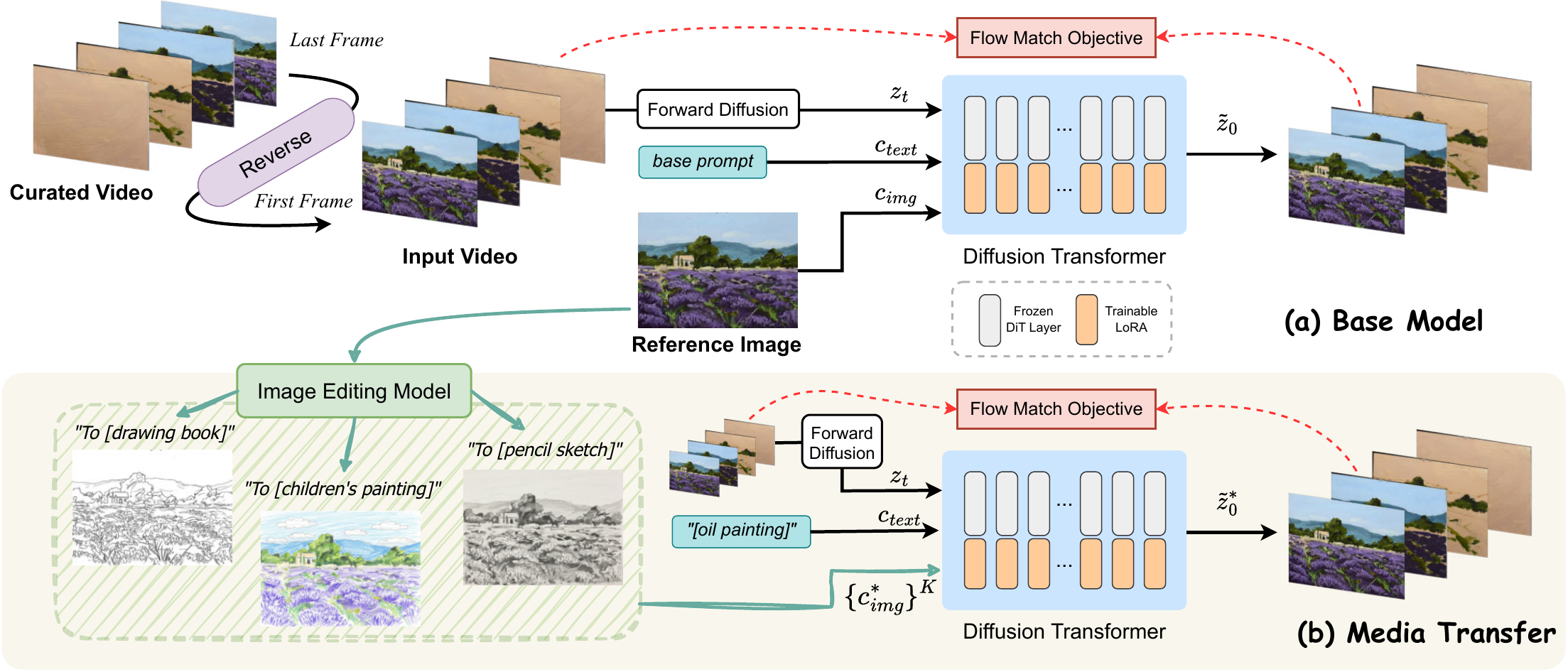}
  \vspace{-6.5mm}
  \caption{Overview of our painting process generation method. The curated video is first reversed to better align with the I2V model. A video generation model conditioned on an input image and a prompt is LoRA-tuned. The input image is a painting, and the model learns to reconstruct the steps to paint it, starting from a finished painting to a blank canvas. In (b), the media transfer model is shown, which enables to render any input image as an acrylic, oil, or pencil painting based on the text input. To achieve this, we generate variations of the reference image using image editing models and train the video generation model to reconstruct the original painting process.}
  \label{fig:method}
  \vspace{-5mm}
\end{figure*}

\noindent \textbf{Neural Painting.}
Neural painting aims to reconstruct visual imagery through a sequence of brushstrokes inferred by a neural network \cite{hu2023stroke, hu2024towards}.
Early efforts, such as Paint Transformer~\cite{liu2021paint}, employed feed-forward architectures based on Transformers~\cite{vaswani2017attention,dosovitskiy2020image} to progressively generate stroke parameters.
Subsequent works \cite{peruzzo2023interactive} also include human interaction in the generation process, similar to~\cite{dall2025collaborative} which lets the user compose an artwork with the help of the neural painter and leverages diversity with a Diffusion Process~\cite{sohl2015deep, ho2020denoising}.
%
Despite their progress, these approaches largely model painting as a parametric rendering problem rather than an authentic artistic process. Synthesized stroke sequences often diverge from how human artists construct compositions. In contrast, our method directly learns from real-world painting videos.

\noindent \textbf{Pixel-Based Generation.}
Pixel-based methods synthesize painting sequences directly at the pixel level, conditioned on a reference image.
Early convolutional methods~\cite{zhao2020painting} attempted to reconstruct painting workflows, while a more recent work Inverse Painting~\cite{chen2024inverse} advanced this painting workflow reconstruction with an autoregressive pipeline. 
More recent diffusion-based approaches further push this paradigm:
ProcessPainter~\cite{song2024processpainter} leverages an image diffusion model to generate the painting process. 
PaintsUndo~\cite{paintsundo} focuses on recreating the painting process for anime-style paintings, leveraging Stable Diffusion~\cite{rombach2022high} as its backbone, while PaintsAlter~\cite{paints_alter} extends this idea to video diffusion; as it has no public implementation at the time of submission, we are unable to include it as a quantitative baseline.
\cite{jiang2025ppjudge} also started to discuss how to evaluate such processes.

\noindent \textbf{Diffusion Models for image and video generation.}
Diffusion Models such as Stable Diffusion~\cite{rombach2022high} and FLUX~\cite{flux2024} have shown great image generation capabilities. A key innovation in these models is the use of a latent space representation, which significantly reduces the computational cost of the diffusion process. For the image generation task a text prompt gets leveraged as guidance. In image-to-image tasks, such as inpainting~\cite{rombach2022high}, an input image serves as an additional conditioning signal.
These concepts have been extended to the video domain~\cite{blattmann2023stable,yang2024cogvideox}. Newer models use the Flow Matching~\cite{lipman2022flow} principle to denoise the latents. Compared to the image generation models, the latent vector contains several frames, and there is a temporal compression that enables coherent multi-frame generation.
Building upon these foundations, pretrained diffusion models can be fine-tuned using various lightweight techniques to adapt them to new concepts, \emph{e.g.}, LoRA \cite{hu2022lora}, ControlNet \cite{zhang2023adding}, and IPAdapter \cite{ye2023ip}, 
all of which are more computationally efficient than training a model from scratch.
%



\section{Methods}
An overview of our method can be seen in Fig. \ref{fig:method}. Our method generates temporally coherent painting sequences that mimic realistic artistic workflows and enable cross-medium transfer. It combines two components: cross-media conditioning, which infuses medium-aware semantics to guide strokes and textures, and reverse-painting learning, which aligns temporal deconstruction with the underlying video generation model. Given a prompt and reference image, our modules guide the model to generate a complete painting process sequence.

\subsection{Video Diffusion}
\label{sec:video_diffusion}
Our method is based on a pretrained Video Diffusion model \cite{wan2025wan} consisting of a video-VAE to encode a given video \(V \in \mathbb{R}^{T \times H \times W \times 3}\) from pixel space into latent space \(\mathbf{x} \in \mathbb{R}^{T/C_{T} \times H / C_{H} \times W / C_{W} \times D}\), where \(C_{T}\), \(C_{H}\) and \(C_{W}\) correspond to a temporal, height, and width compression ratio respectively. Typically \cite{wan2025wan, yang2024cogvideox, hacohen2024ltx, peng2025open}, the temporal compression \(C_{T}\) is set to 4 or 8, while the spatial ratios \(C_{H}\),\(C_{W}\) are 8 or higher, resulting in a highly compact latent representation. This latent is used to train a Diffusion Transformer (DiT) \cite{peebles2023scalable} with the Flow Matching objective \cite{lipman2022flow, esser2024scaling}. Given a video latent \(\mathbf{x_{1}}\), random noise \( \mathbf{x_0} \sim \mathcal{N}(0, \mathbf{I}) \) and timestep \( t \in [0, 1] \), a linear interpolation \(\mathbf{x_t} = t \mathbf{x_1} + (1-t)\mathbf{x_0}\) can be defined. The target velocity can be defined as \(\mathbf{v_t = x_1 - x_0}\). By modeling the DiT output as a prediction of this velocity vector, one can formulate the loss function as the mean squared error (MSE) between the model output and \(v_t\),

\begin{equation}
    \mathcal{L} = \mathbb{E}_{\mathbf{x}_0, \mathbf{x}_1, \mathbf{c}_{\text{text}}, t} \left\| u(\mathbf{x}_t, \mathbf{c}_{\text{text}}, t; \theta) - \mathbf{v}_t \right\|_2^2,
\end{equation}
\( \mathbf{c}_{\text{text}} \) is the text embedding sequence, \( \theta \) denotes the model parameters, and \( u(\mathbf{x}_t, \mathbf{c}_{\text{text}}, t; \theta) \) is the predicted velocity.

\noindent Building on text-to-video models, image-to-video (I2V) approaches \cite{wan2025wan, lin2024open, peng2025open} take an input image \(I \in \mathbb{R}^{H \times W \times 3}\) and encode it into the latent space to serve as the initial frame. Depending on the architecture \cite{wan2025wan}, this is achieved either by prepending it directly to the noise latent or by concatenating it along the channel axis. The combined latent representation is then processed by the DiT while the encoded input latent remains fixed throughout the generation.


\begin{figure}
    \centering
    \includegraphics[width=\linewidth]{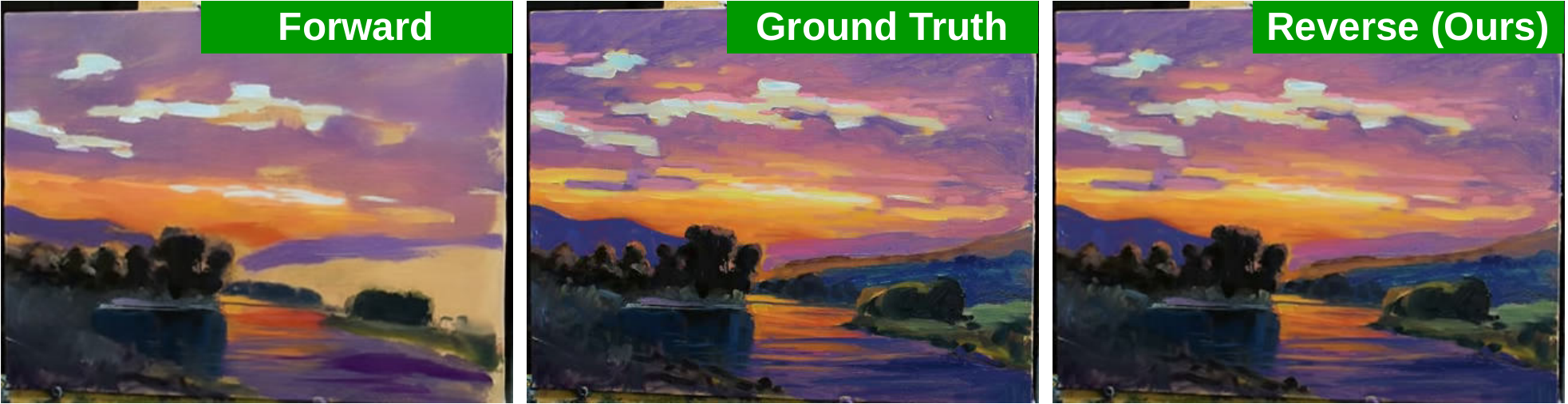}
    \vspace{-7mm}
    \caption{Comparison of final frame. On the left, the forward process is not able to reconstruct the input image fully; details are still missing, i.e., bottom right part. Image courtesy Samir Godinjak, from the Painting with Samir YouTube channel.}
    \label{fig:ablation}
    \vspace{-3mm}
\end{figure}

\begin{figure}
    \centering
    \includegraphics[width=\linewidth]{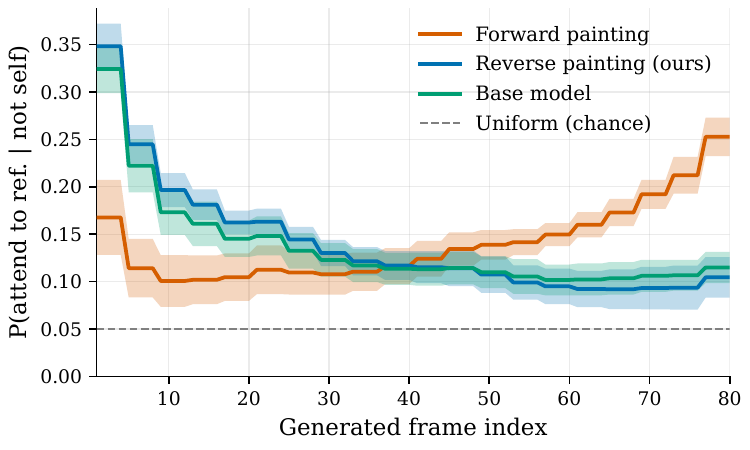}
    \vspace{-7mm}
    \caption{Reference attention distribution (WAN I2V 5B). Probability of attending to the reference frame relative to non-self frames ($P(\text{attend to ref.} \mid \text{not self})$). Our reverse painting model maintains a decaying attention profile similar to the base model, whereas forward painting shifts focus heavily toward later frames. Fine-tuned models were trained for 7 epochs. Stepwise plateaus of 4 frames reflect the model's $4\times$ temporal latent compression. See \cref{fig:appx:attn_vis_figure} in the appendix for additional architectures.}
    \label{fig:attn_vis_figure}
    \vspace{-4mm}
\end{figure}

\begin{figure*}[!t]
    \centering
    \includegraphics[width=\linewidth]{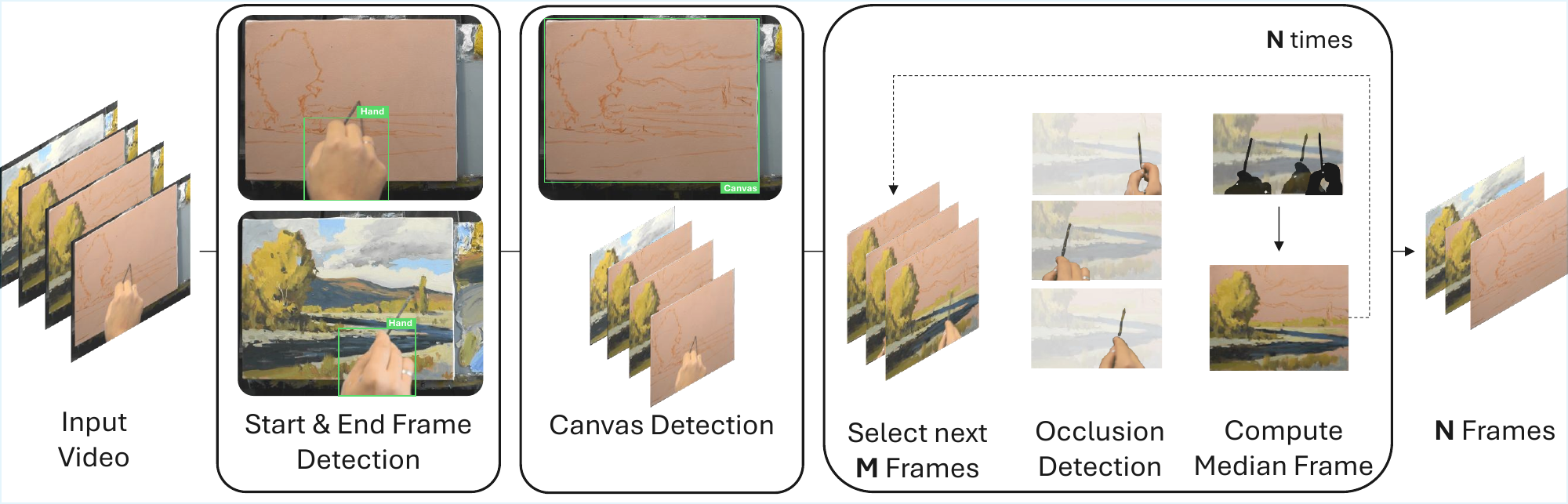}
    \vspace{-7mm}
    \caption{Dataset Curation Pipeline Overview. Our framework extracts painting workflows from raw tutorial videos. First, start and end frames are detected, and the painting canvas is localized. The video is then partitioned into $N$ segments, from which $M$ frames are sampled per segment. Subsequently, occlusions (e.g., hands, brushes) are detected in the sampled frames, and a masked median is computed over the sampled frames, using the preceding median frame as a reference to remove transient obstructions. Producing $N$ occlusion-free frames. Image courtesy of Samir Godinjak, from the Painting with Samir YouTube channel.}
    \label{fig:pipe}
    \vspace{-5mm}
\end{figure*}

\subsection{Art Media Transfer}
\label{sec:semantic}

\noindent A central challenge in artistic process generation is reproducing the distinct procedural evolution of different media (\emph{e.g.}, color layering in oils vs. hatching in sketches) from arbitrary input images. To enable transfer across artistic media, we train the video generation model using a many-to-one mapping strategy: we generate multiple style-transformed variants of a reference artwork, then train the model to reconstruct the original painting process guided by the target medium prompt.

\noindent Specifically, given a painting video $V$ with a final frame $I$, a textual medium description $m$, and a scene description $s$, we construct the combined semantic prompt $p = [m; s]$ and its text embedding $\mathbf{c}_{\text{text}}$. We apply an off-the-shelf style transformation to $I$ to obtain $I^*$, preserving the underlying scene geometry and semantics while stripping away the original artistic medium's identity. 

\noindent To enforce media-aware synthesis, we replace the standard image condition $I$ in the I2V Flow Matching objective with its stylized counterpart $I^*$, while keeping the target video latent $\mathbf{x}_1$ unchanged:
\begin{equation}
    \mathcal{L}_{\textit{I2V}} = \mathbb{E}_{\mathbf{x}_0, \mathbf{x}_1, I^*, \mathbf{c}_{\text{text}}, t} \left\| u(\mathbf{x}_t, I^*, \mathbf{c}_{\text{text}}, t; \theta) - \mathbf{v}_t \right\|_2^2.
\end{equation}

\noindent By decoupling the input image style from the target execution process, this formulation forces the model to ignore the source medium's texture. Instead, it extracts structural cues (\emph{e.g.}, contours and spatial layouts) from $I^*$ and maps them directly to the procedural stroke dynamics and rendering trajectory specified by $\mathbf{c}_{\text{text}}$.

\begin{table}
    \centering
    \scriptsize
    \begin{tabular}{lccccc}
    \toprule
     & \textbf{Acrylic} & \textbf{Oil} & \textbf{Pencil} & \textbf{Loomis} & \textbf{Total} \\
    \midrule
    \textbf{\# of Videos} & 81 & 151 & 298 & 207 & \textbf{737} \\
    \textbf{Avg. Duration [min]} & $\approx$40 & $\approx$30 & $\approx$12 & $\approx$20 & --- \\
    \bottomrule
    \vspace{-5mm}
    \end{tabular}
    \caption{Overview of the curated video dataset by art medium.}
    \label{tab:dataset}
    \vspace{-4mm}
\end{table}

\subsection{Reverse-Painting Learning Strategy}
\label{sec:reverse}

When fine-tuning the I2V model on the natural (forward) painting order, the model consistently failed to reconstruct the finished artwork in the final frame, with several painting steps still missing (\cref{fig:ablation}). Since video diffusion models primarily attend to nearby frames \cite{wan2025wan, radial_attn}, generating the final painting requires maintaining long-range dependencies to the reference image at the first frame, substantially altering the pretrained attention pattern. \cref{fig:attn_vis_figure} confirms this.

\noindent To preserve this temporal prior, we train on reversed painting sequences. Given a painting process $V=\{f_1,\ldots,f_T\}$, we train on the reversed sequence
$V_{\mathrm{rev}}=\{f_T,\ldots,f_1\}$.
The model therefore learns to progressively remove details rather than add them, relying on local temporal consistency instead of persistent reference attention. After inference, the generated sequence is simply reversed back to a forward painting order. This lets the model faithfully reconstruct the reference (\cref{fig:ablation}), with an attention pattern matching the base model's (\cref{fig:attn_vis_figure}).


\section{Dataset Curation Pipeline}
\label{sec:dataset}
An overview of our pipeline is shown in Fig. \ref{fig:pipe}. Our process begins with temporal trimming, where we detect the first and last appearance of a "hand" using GroundingDINO \cite{liu2024grounding} to isolate the core painting process. Next, for canvas localization, we first attempt to find the "canvas" using GroundingDINO; for split-screen tutorials (\emph{e.g.}, Loomis), we instead compute the maximum horizontal intensity gradient to separate the reference photo from the canvas. We then partition the video into 10-second segments, sampling 30 frames (3 fps) from each. To segment occlusions (\emph{e.g.}, hands, brushes), we employ InSPyReNet \cite{kim2022revisiting} for pencil portraits and BiRefNet \cite{BiRefNet} for other domains; this hybrid approach was empirically found to yield superior foreground masks across diverse styles. A clean frame for each segment is generated by computing a masked median of its samples; this calculation iteratively incorporates prior frames to fill persistent occlusions.


\noindent \textbf{Data Collecting. }
We curated a dataset from painting tutorial videos on YouTube, prioritizing static camera angles and minimal canvas movement. All videos were processed with the pipeline described in Section \ref{sec:dataset} and manually reviewed to filter out poor-quality results. A detailed breakdown of our final dataset is provided in Table \ref{tab:dataset}.


\noindent \textbf{Fine-Tuning Datasets. }
\label{sec:fine-tuning-dataset}
We merge the Acrylic, Loomis Portrait, Oil, and Pencil subsets into a unified dataset for fine-tuning. Each entry includes a reference frame (the finished artwork) and its progressive painting states. We will release the configuration files including: (1) Public source URLs, start/end frame indices, and canvas bounding boxes. (2) computed pixel-level occlusion masks. (3) The pipeline code that downloads, crops, and removes occlusions in the sequences. This approach adheres to licensing constraints while ensuring reproducibility of the dataset.
\begin{figure}
    \centering
    \includegraphics[width=\linewidth]{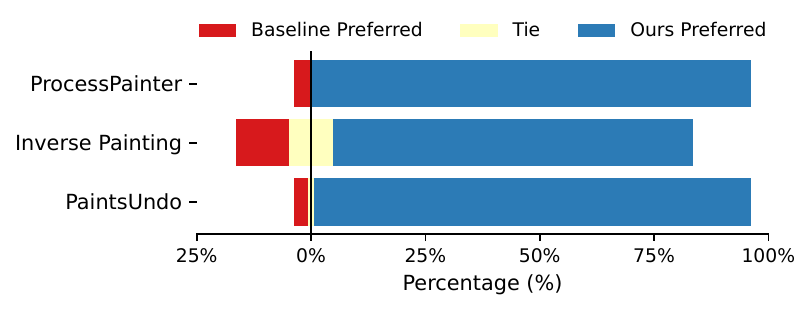}  
    \vspace{-7mm}
    \caption{User preference evaluation (N=22). Blue bars indicate a preference for our method, red bars indicate a preference for the baseline, and the yellow center represents a tie. Users preferred our results over ProcessPainter, Inverse Painting, and PaintsUndo in 96.2\%, 78.7\%, and 95.6\% of comparisons, respectively.}
    \label{fig:user_preferences}
    \vspace{-4mm}
\end{figure}

\begin{figure*}[htbp!]
    \centering
   
    \begin{subfigure}[b]{\textwidth}
        \centering
        \includegraphics[width=\textwidth]{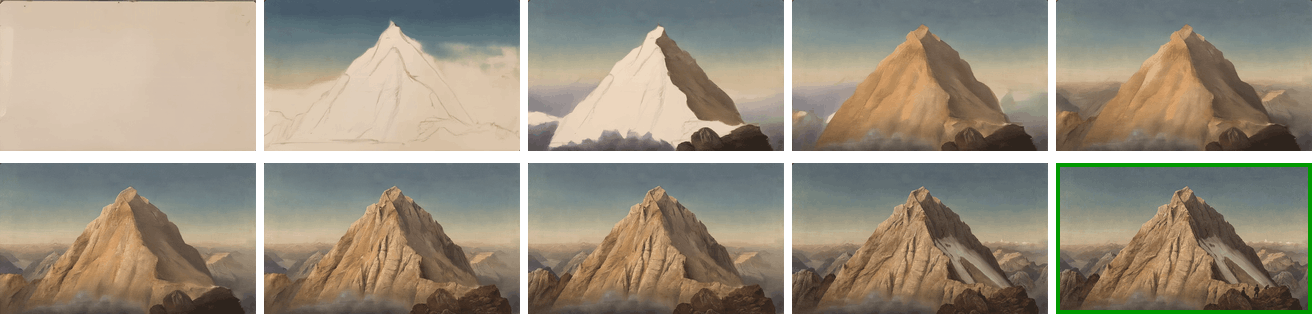}
        \vspace{-5mm}
        \caption{Reconstruction of the Painting process. \textit{Peak of Mount Triglav} by Marko Pernhart, Public Domain, via Wikimedia Commons.}
    \end{subfigure}

    \vspace{2mm}
    \begin{subfigure}[b]{\textwidth}
        \centering
        \includegraphics[width=\textwidth]{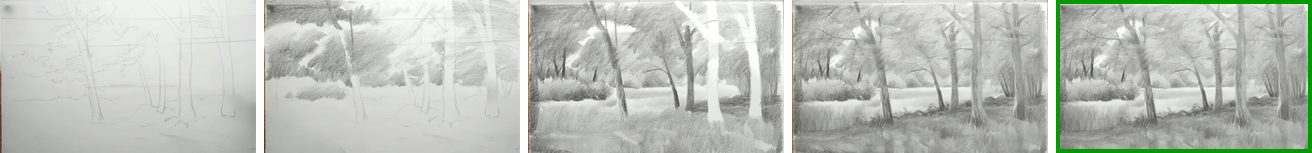}
        \vspace{-5mm}
        \caption{Reconstruction of the Painting process. Input image is generated.}
    \end{subfigure}

    \vspace{2mm}
    \begin{subfigure}[b]{\textwidth}
        \centering
        \includegraphics[width=\textwidth]{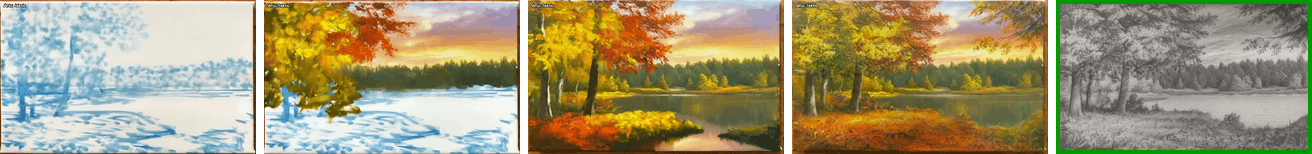}
        \vspace{-5mm}
        \caption{Art media transfer, pencil sketch to acrylic painting. Input image is a style-transferred version from \cite{aluart_autumn} via ControlNet.}
    \end{subfigure}

    \vspace{2mm}
    \begin{subfigure}[b]{\textwidth}
        \centering
        \includegraphics[width=\textwidth]{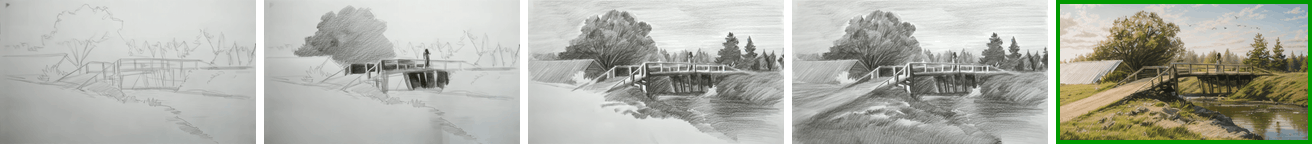}
        \vspace{-5mm}
        \caption{Art media transfer, color painting to pencil sketch. Input image is a style-transferred version from \cite{paintlane2025bridge} via ControlNet.}
    \end{subfigure}

    \vspace{2mm}
    \begin{subfigure}[b]{\textwidth}
        \centering
        \includegraphics[width=\textwidth]{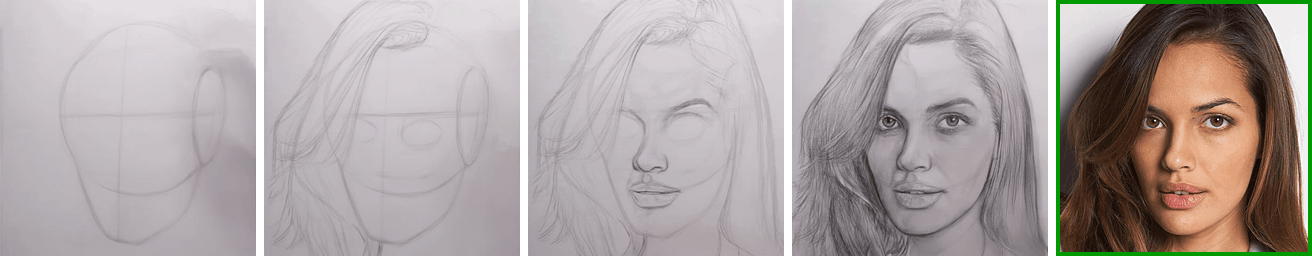}
        \vspace{-5mm}
        \caption{Art media transfer, from portrait photograph to loomis style pencil sketch. Image from CelebA-HQ \cite{karras2017progressive}}
    \end{subfigure}

    \vspace{-2mm}
    \caption{Sampled frames of our method with input images. Painting order from left to right, top to bottom. The image with the green border is the reference painting. (a) Illustrates a typical painting process where the model first creates a sketch of the mountain, paints the background, layers the colors of the foreground objects, and finishes with fine details. (b) and (d) Show pencil paintings where outlines are sketched first, followed by details and depth added via hatching. (c) An acrylic painting process where the scene gets first sketched with blue color before being painted, the input image is a pencil sketch. (e) Demonstrates the Loomis portrait method, which uses simple circles and division lines to faithfully recreate the head.}
    \label{fig:showcase}
    \vspace{-3mm}
\end{figure*}

\begin{table*}[htpb!]
  \centering
  \setlength{\tabcolsep}{8pt}
  \vspace{-2mm}
  \begin{tabular}{l|ccccc|cc|r}
    \toprule
    \textbf{Method} & \textbf{FID $\downarrow$} & \textbf{LPIPS $\downarrow$} & \textbf{CLIP $\uparrow$} & \textbf{DINOv2 $\uparrow$} & \textbf{DDC $\downarrow$} & \textbf{Latency (s)} & \textbf{FPS} & \textbf{Params} \\

    \midrule
    Inverse Painting ~\cite{chen2024inverse} & 326.15 & 0.61 & 0.66 & 0.21 & 48.77 & 333.8 & 0.16 & 900M \\
    ProcessPainter~\cite{song2024processpainter} & 282.90 & 0.53 & 0.76 & 0.50 & 17.79 & 1.7 & 4.71 & 1.3B \\
    PaintsUndo~\cite{paintsundo} & 236.52 & 0.55 & 0.77 & 0.56 & 15.05 & 129.6 & 0.62 & 900M \\
    PaintsUndo$^\dagger$~\cite{paintsundo} & \underline{196.66} & \underline{0.43} & \underline{0.83} & \underline{0.66} & \underline{10.02} & 129.6 & 0.62 & 900M\\
    \midrule
    Ours (VACE) & 183.61 & 0.42 & 0.83 & 0.70 & 9.18 & 171.7 & 0.47 & 1.3B \\
    Ours (Wan 2.2) & 152.42 & \textbf{0.38} & 0.85 & 0.75 & \textbf{7.97} & 37.3 & 2.17 & 5B \\
    Ours (Wan 2.1) & \textbf{151.04} & \textbf{0.38} & \textbf{0.86} & \textbf{0.76} & 8.39 & 428.3 & 0.19 & 14B \\
    \bottomrule
  \end{tabular}
  \vspace{-2mm}
  \caption{Evaluation using various metrics for measuring fidelity and inference speed. The number of parameters only relates to the image/video generation component. $^\dagger$ denotes fine-tuned on our dataset. Best overall results in bold, best baseline results underlined.}
  \label{tab:art_evaluation}
  \vspace{-2mm}
\end{table*}

\begin{table}[ht!]
    \vspace{-5pt}
    \scriptsize
    \renewcommand{\arraystretch}{0.9}
    \setlength{\tabcolsep}{2.8pt}
    \centering
    
    \begin{minipage}[c]{0.78\columnwidth}
    \centering
    
    \begin{tabular}{lcccccc|c}
    \toprule
    \textbf{Method $|$ Video} & \textbf{1} & \textbf{2} & \textbf{3} & \textbf{4} & \textbf{5} & \textbf{6} & \textbf{Avg.} \\
    \midrule
    \multicolumn{8}{l}{\textit{Stroke Coherence}} \\
    \quad PaintsUndo & $2.4$ & $1.5$ & $1.6$ & $2.0$ & $1.6$ & $1.0$ & $1.7_{\pm 1.1}$ \\
    \quad Ours & $3.4$ & $4.2$ & $3.4$ & $4.0$ & $3.8$ & $4.3$ & $\underline{3.9_{\pm 1.0}}$ \\
    \quad GT & $4.2$ & $4.8$ & $4.8$ & $4.4$ & $4.0$ & $5.0$ & $\mathbf{4.5_{\pm 0.9}}$ \\
    \midrule
    \multicolumn{8}{l}{\textit{Layering Logic}} \\
    \quad PaintsUndo & $2.2$ & $1.5$ & $1.1$ & $2.2$ & $1.6$ & $1.1$ & $1.7_{\pm 1.1}$ \\
    \quad Ours & $3.5$ & $4.8$ & $3.8$ & $4.4$ & $3.8$ & $4.4$ & $\underline{4.1_{\pm 1.0}}$ \\
    \quad GT & $4.4$ & $4.6$ & $5.0$ & $4.2$ & $4.2$ & $5.0$ & $\mathbf{4.6_{\pm 0.8}}$ \\
    \midrule
    \multicolumn{8}{l}{\textit{Color Consistency}} \\
    \quad PaintsUndo & $3.3$ & - & $2.0$ & - & - & $1.4$ & $2.3_{\pm 1.2}$ \\
    \quad Ours & $4.0$ & - & $3.6$ & - & - & $4.3$ & $\underline{4.0_{\pm 0.9}}$ \\
    \quad GT & $4.2$ & - & $4.8$ & - & - & $5.0$ & $\mathbf{4.7_{\pm 0.7}}$ \\
    \midrule
    \multicolumn{8}{l}{\textit{Overall Realism}} \\
    \quad PaintsUndo & $2.1$ & $1.2$ & $1.1$ & $2.1$ & $1.4$ & $1.2$ & $1.6_{\pm 1.0}$ \\
    \quad Ours & $2.9$ & $4.3$ & $3.2$ & $3.8$ & $3.8$ & $4.2$ & $\underline{3.7_{\pm 1.2}}$ \\
    \quad GT & $3.9$ & $4.1$ & $4.8$ & $4.0$ & $4.0$ & $5.0$ & $\mathbf{4.3_{\pm 1.1}}$ \\
    \midrule
    \multicolumn{8}{l}{\textit{Educational Usefulness}} \\
    \quad PaintsUndo & $1.8$ & $1.1$ & $1.1$ & $1.9$ & $1.2$ & $1.1$ & $1.4_{\pm 1.0}$ \\
    \quad Ours & $2.6$ & $4.2$ & $3.2$ & $3.9$ & $3.5$ & $4.0$ & $\underline{3.6_{\pm 1.2}}$ \\
    \quad GT & $2.9$ & $4.5$ & $4.6$ & $3.6$ & $3.8$ & $4.8$ & $\mathbf{4.0_{\pm 1.1}}$ \\
    \bottomrule
    \end{tabular}
    
    \end{minipage}
    \hfill
    \begin{minipage}[r]{0.18\columnwidth}
        \centering
        \includegraphics[width=0.98\linewidth]{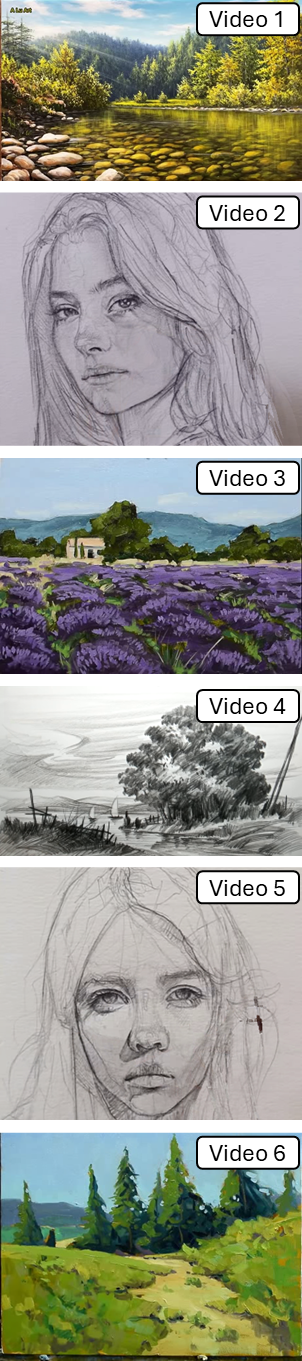} 
    \end{minipage}

    \vspace{-5pt}
    \caption{Expert evaluation results. Participants rated videos across five metrics on a 5-point Likert scale. Our method consistently outperforms the baseline, approaching Ground Truth (GT) in several criteria.}
    \label{tab:expert_study}
    \vspace{-3mm}
\end{table}


\section{Experiments}
\label{sec:experiments}
\noindent\textbf{Implementation Details.}
We base our method on Wan 2.1 14B 480p I2V \cite{wan2025wan}. Its uncompressed first frame aligns perfectly with our reverse-painting strategy to accurately reconstruct the input image. Images are resized and padded to 480 $\times$ 832. Using the dataset from \cref{sec:fine-tuning-dataset}, we train LoRA adapters for 14 epochs (empirically optimal) with a learning rate of $1e-4$, taking 24 hours on 4 NVIDIA H100 GPUs. To demonstrate scalability, we apply identical hyperparameters to fine-tune smaller models (Wan 2.2 5B and VACE 1.3B \cite{jiang2025vace}) on a single NVIDIA A100 (64GB) GPU. Datasets are split 80/10/10 for training, validation, and testing.


\noindent For the media transfer dataset, we generate style-transformed images \(I^*\) using ControlNet (for sketches and Loomis portraits) and Qwen Image Edit \cite{wu2025qwen} (for acrylic and oil); more details in the appendix \cref{appx:painting_media_transfer_dataset}.
To enable media transfer, we initialize a separate LoRA from our 7-epoch base model and train it for 7 additional epochs. 


\subsection{Baselines}
We compare our approach against three representative methods:
Inverse Painting \cite{chen2024inverse}, which autoregressively reconstructs the painting process by determining which area in the painting should be filled next and inpainting the selected area with a diffusion model.
ProcessPainter \cite{song2024processpainter}, which learns to reconstruct the painting process using only a few real painting examples.
PaintsUndo \cite{paintsundo}, designed for reconstructing the painting process with fine-grained control over the painting progression. 

\subsection{Quantitative Comparison}
We evaluate our method against the baselines using LPIPS, CLIP and Dinov2 (to assess perceptual similarity to human workflows) and FID \cite{DOWSON1982450} (to measure distributional alignment with ground-truth painting sequences). For each method, generated frames are compared to the closest corresponding ground-truth frame in terms of painting progression. Metrics are first averaged across frames within each test video, and then averaged across all test videos. This approach ensures that poor performance on individual videos is appropriately reflected in the overall evaluation.
To evaluate convergence speed of the painting process we report DDC introduced in \cite{chen2024inverse}.

\noindent The results are shown in ~\cref{tab:art_evaluation}, our proposed method achieves state-of-the-art performance across all evaluated metrics. Notably, our 5B model achieves comparable performance to the 14B model. Even our smallest 1.3B model maintains robust generation capabilities, surpassing the best fine-tuned baseline in both fidelity and temporal consistency.
Inference efficiency was measured on a single NVIDIA A100 GPU. The 5B model achieves superior throughput compared to the 14B and 1.3B (VACE) models due to its more compressed latent space and reduced token count. With a peak VRAM footprint of 35.3~GB, the 5B model consumes more memory than the 1.3B model (18.4~GB) but less than the 14B model (47.0~GB), striking the most effective trade-off.

\begin{table}[t]
    \vspace{-5pt}
    \centering
    \small
    \begin{tabular}{cc|cc|cc}
    \hline
    \multicolumn{2}{c|}{} & \multicolumn{2}{c|}{\textbf{Forward}} & \multicolumn{2}{c}{\textbf{Reverse (Ours)}} \\
    \textbf{Model} & \textbf{Epoch} & \textbf{FID}$\downarrow$ & \textbf{LPIPS}$\downarrow$ & \textbf{FID}$\downarrow$ & \textbf{LPIPS}$\downarrow$ \\
    \hline
    5B & 1 & 171 & 0.414 & 187 & 0.436 \\
    5B & 3 & 179 & 0.424 & 166 & 0.395 \\
    5B & 5 & 175 & 0.424 & 160 & \textbf{0.389} \\
    5B & 7 & 163 & 0.401 & \textbf{159} & 0.390 \\
    \hline
    14B & 7 & 173 & 0.417 & \textbf{164} & \textbf{0.393} \\
    \hline
    \end{tabular}
    \vspace{-2mm}
    \caption{Ablation study evaluating the impact of our proposed reverse painting training strategy. Comparing the base video generation model against our method under the same training constraints demonstrates that the reverse painting strategy converges faster and improves visual fidelity.}
    \label{tab:ablation}
\end{table}

\begin{figure*}[htbp!]
    \centering
    \begin{subfigure}[b]{0.49\linewidth}
        \centering
        \includegraphics[width=\linewidth]{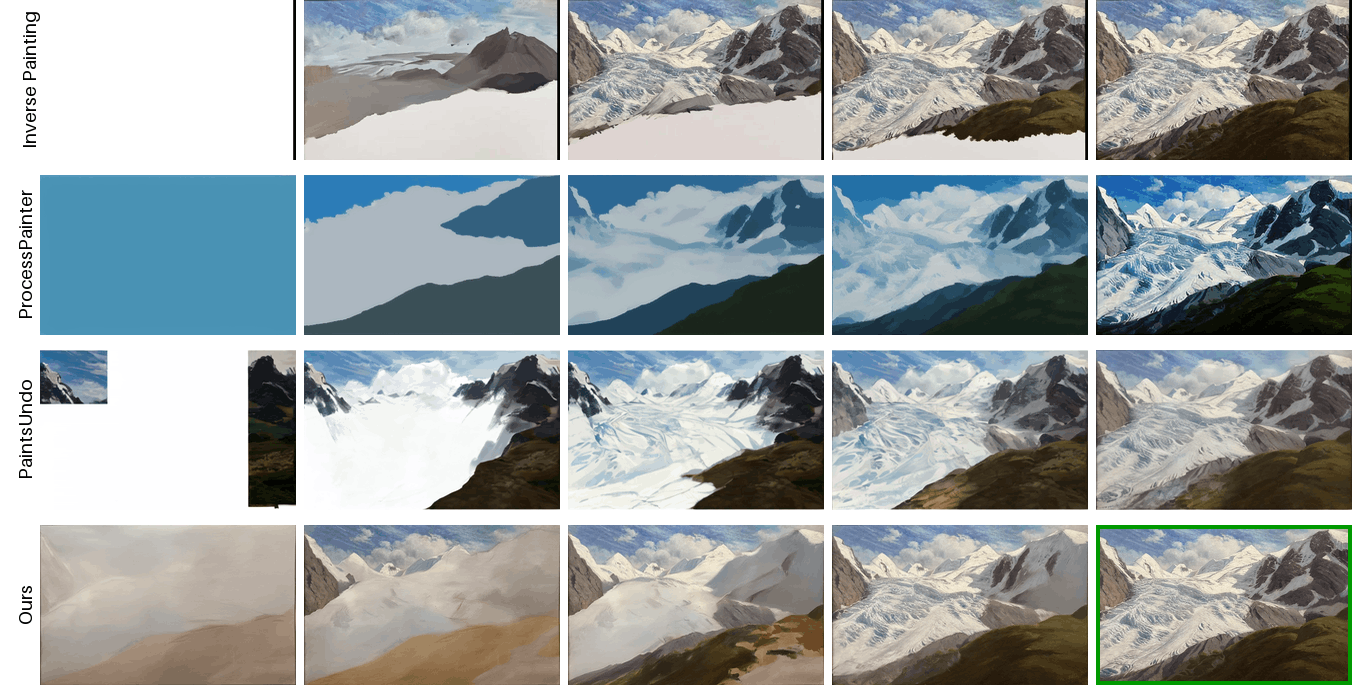}
        \caption{\scriptsize \textit{Glacier d'Arolla} by François Diday, Public Domain, via Wikimedia Commons.}
    \end{subfigure}
    \hfill
    \begin{subfigure}[b]{0.49\linewidth}
        \centering
        \includegraphics[width=\linewidth]{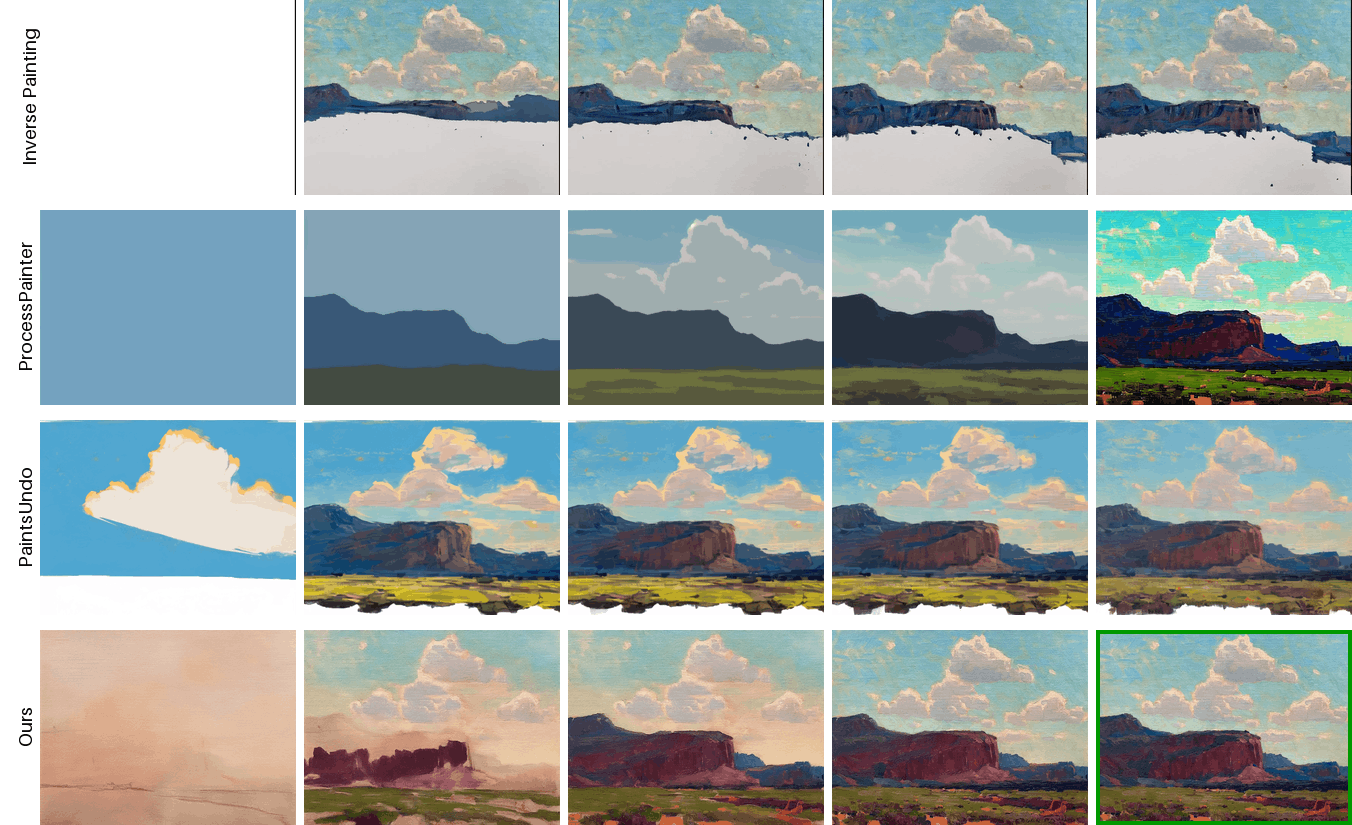}
        \caption{\scriptsize \textit{Continental Divide} by Edgar A. Payne, Public Domain, via Wikimedia Commons.}
    \end{subfigure}
    \vspace{-2mm}
    \caption{Method comparison. Painting order from left to right. The final frame (green border) is visually nearly indistinguish-
able from the reference image; the latter is omitted for brevity.}
    \label{fig:art_media_comparison_000}
\end{figure*}

\subsection{Human Evaluation} 
\noindent \textbf{Expert Evaluation.}
To assess whether our method produces realistic and human-aligned painting sequences, we recruited 17 professional artists, art students, and educators. Participants evaluated 30-second videos of painting processes generated by PaintsUndo, our method, and the Ground Truth (GT) across six representative artworks. To ensure an unbiased assessment, the videos were evaluated independently without direct side-by-side comparisons. The experts rated each video on a 5-point Likert scale (1 = Strongly Disagree, 5 = Strongly Agree) across five process-specific criteria (see \cref{tab:expert_study}).  See \cref{sec:appx:human_eval} for implementation details.

\noindent Key Findings: \textbf{(1) Realistic Artistic Decisions}: Our method outperforms PaintsUndo across all metrics. For pencil and Loomis sketches (Video 2, 4 \& 5) and simple landscape paintings (Video 6), experts rated our stroke coherence, layering logic, and color consistency as highly natural, closely approaching GT.

\noindent \textbf{(2) Educational Value}: Experts noted that compressing complex painting processes (Video 1) into 30 seconds reduces pedagogical value for both GT and our method equally. However, simpler examples (Videos 2, 4-6) demonstrate our method's educational potential.

\noindent \textbf{Human Preferences.}
To evaluate general perceptual preference at scale, we conducted a user study across 25 in-the-wild paintings with 22 non-expert participants. As shown in \cref{fig:user_preferences}, our method significantly outperforms all baselines ($p < 0.001$), achieving win rates of 96.2\% against ProcessPainter, 78.7\% against Inverse Painting, and 95.6\% against PaintsUndo. See \cref{sec:appx:human_eval} for implementation details.


\subsection{Ablation}
To demonstrate that the un-painting (reverse) frame order during LoRA tuning yields better results and converges faster we fine-tuned the video generation model on both frame ordering. Quantitative results can be seen in \cref{tab:ablation} where Forward indicates the painting order as seen in painting tutorial videos, from blank canvas to finished painting. Reverse (Ours) indicates the reversed frame ordering that we used in our method. The forward method converges slower since it needs to drastically change the attention pattern as seen in \cref{fig:attn_vis_figure}, while having worse visual results than our proposed method.
In the appendix we conducted further experiments to verify the art media transfer model, see \cref{sec:appx:art_media_transfer}. 

\subsection{Qualitative Results}
Example painting processes are shown in \cref{fig:teaser,fig:showcase}. Our model reconstructs diverse paintings by first establishing the global structure before progressively adding color, shading, and fine details. Pencil sketches begin with outlines and hatching, while color paintings typically follow a background-to-foreground layering strategy consistent with traditional painting practice \cite{carlson1973guide, albala2009landscape}.

\noindent Our cross-media model preserves scene structure while adapting both the appearance and painting procedure to the target medium, enabling transformations such as sketches to paintings, paintings to sketches, and Loomis-style portrait construction. Additional examples are provided in the supplementary material and appendix.





\subsection{Qualitative Comparison}

Comparison of the methods can be seen in \cref{fig:art_media_comparison_000}. Inverse Painting follows a fixed background-to-foreground order and is less flexible than our method; it also sometimes fails to fully converge (\cref{fig:art_media_comparison_000}b). ProcessPainter is limited to eight generated frames, yielding a coarse process with synthetic-looking, unnatural stroke steps. PaintsUndo produces more coherent sequences but tends to progress too quickly in early stages, with some frame-to-frame inconsistency. Our method most closely mirrors the ground truth, progressing from broad background colors to layered detail in a way that aligns with human painting workflows.





\section{Conclusion}
\label{sec:conclusion}
We presented Loomis Painter, a framework for reconstructing and transferring artistic painting processes with video diffusion models. Our key insight is that training on reversed painting sequences better aligns with the temporal inductive bias of pretrained image-to-video models, preserving their attention patterns and enabling faithful reconstruction of the final artwork. Combined with our cross-media training strategy, the model generates human-aligned painting processes across diverse artistic media while supporting art-media transfer from input images.

\noindent To support this task, we introduced a pipeline to process a large-scale dataset of traditional painting videos. Extensive quantitative, qualitative, and human evaluations demonstrate state-of-the-art performance over existing methods. We hope this work enables personalized painting tutorials, educational tools, and future research on procedural artistic video generation.




\section{Limitation \& Future Work}
\label{sec:limitation}

\noindent\textbf{Limitations.}
There exist some artifacts in our dataset. Our occlusion detector in the data collection pipeline cannot detect hand shadows, leading to dark artifacts in the training data. These artifacts are mainly visible in pencil painting generations, typically in the bottom-right region.
%

\noindent\textbf{Future Work.}
To fully support human artists in their painting journey, simply showing a step-by-step sequence is not enough. Understanding the painting process also requires indicating which colors were selected, how they were mixed, which tools (pencils or brushes) were used, and how to apply them on the canvas.

\section{Acknowledgement}
This work was supported by the EU Horizon project “ELIAS - European Lighthouse of AI for Sustainability” (No. 101120237) and the FIS project GUIDANCE (Debugging Computer Vision Models via Controlled Cross-modal Generation) (No. FIS2023-03251).

\noindent We acknowledge ISCRA for awarding this project access to the LEONARDO \cite{leonardo} supercomputer, owned by the EuroHPC Joint Undertaking, hosted by CINECA (Italy) .
{
    \small
    \bibliographystyle{ieeenat_fullname}
    \bibliography{main}
}

\clearpage

\section{Experiments}
\subsection{Qualitative Results}
\begin{figure*}[htbp!]
    \centering

    \begin{subfigure}[b]{\textwidth}
        \centering
        \includegraphics[width=\textwidth]{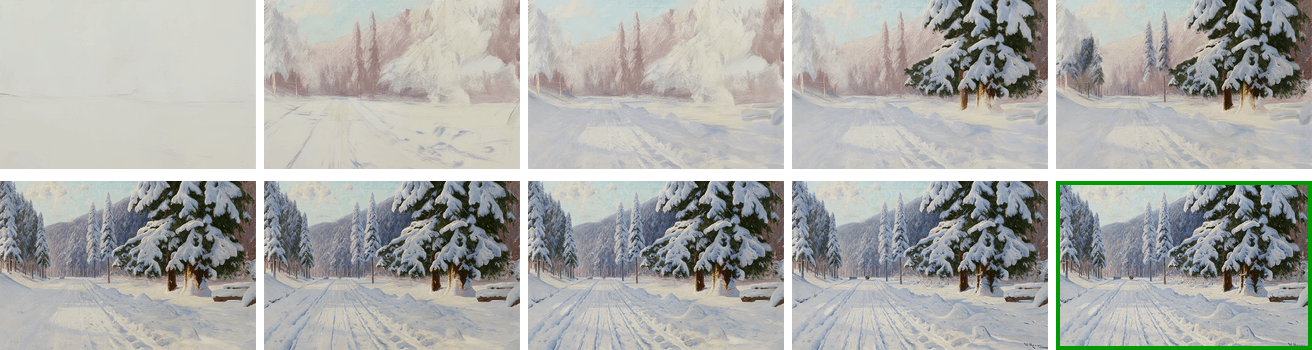}
        \vspace{-5mm}
        \caption{\textit{Sonniger Wintertag im verschneiten Wald} by Walter Moras, Public Domain, via Wikimedia Commons.}
    \end{subfigure}
    
    \vspace{2mm}
    \begin{subfigure}[b]{\textwidth}
        \centering
        \includegraphics[width=\textwidth]{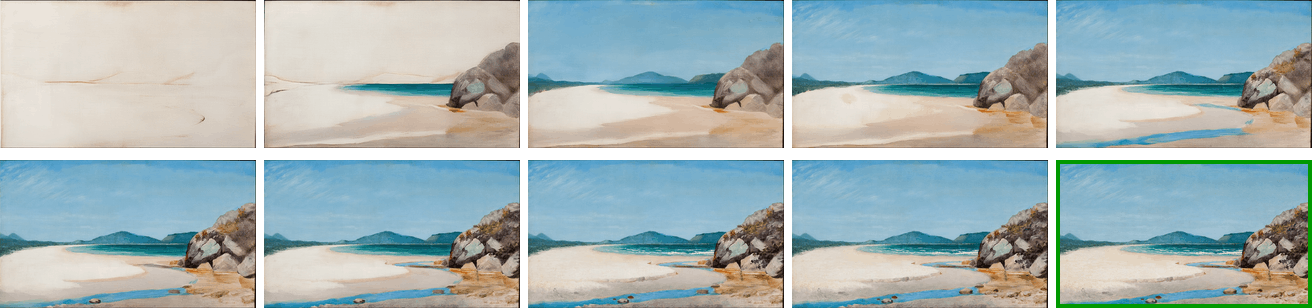}
        \vspace{-5mm}
        \caption{\textit{Seascape, Guarujá} by Almeida Júnior, Public Domain, via Wikimedia Commons.}
    \end{subfigure}

    \vspace{2mm}
    \begin{subfigure}[b]{\textwidth}
        \centering
        \includegraphics[width=\textwidth]{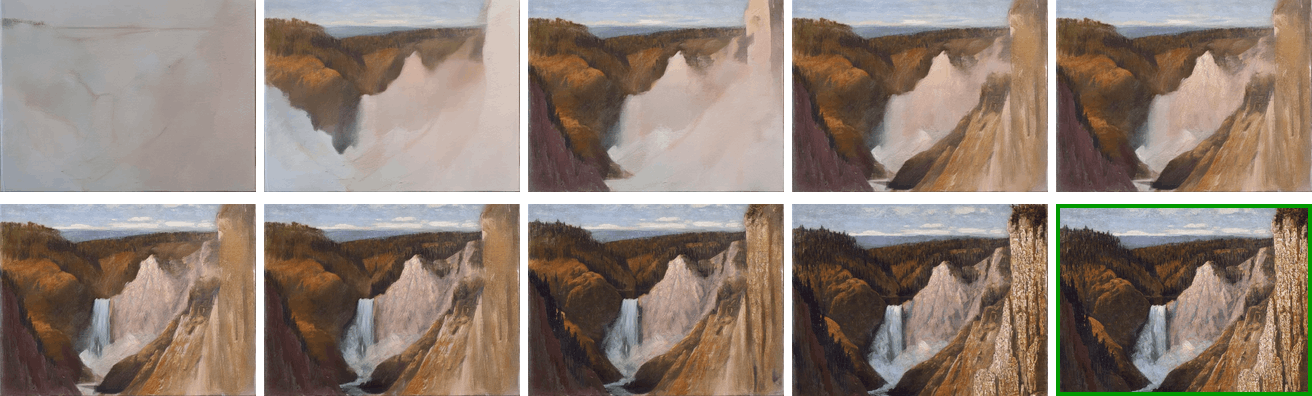}
        \vspace{-5mm}
        \caption{\textit{View of the Lower Falls, Grand Canyon of the Yellowstone} by Grafton Tyler Brown, Public Domain, via Wikimedia Commons.}
    \end{subfigure}

    \vspace{2mm}
    \begin{subfigure}[b]{\textwidth}
        \centering
        \includegraphics[width=\textwidth]{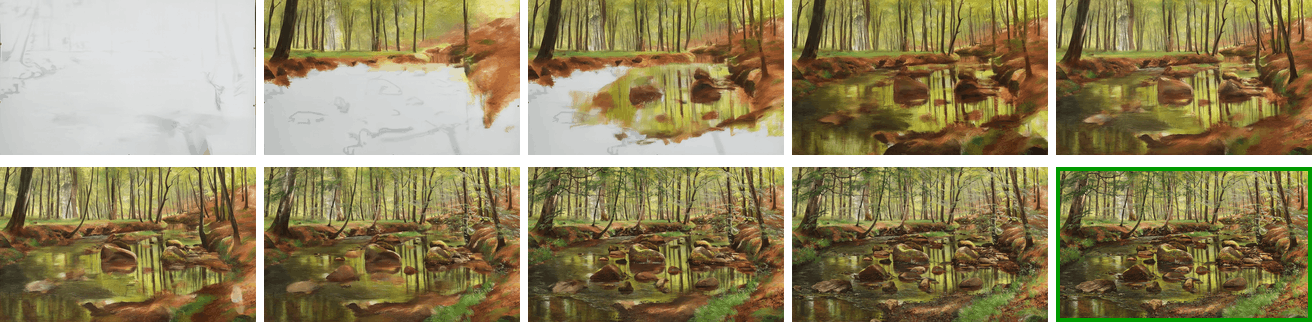}
        \vspace{-5mm}
        \caption{\textit{Forårsdag i skoven med solglimt i åen} by Carl Frederik Aagaard, Public Domain, via Wikimedia Commons.}
    \end{subfigure}

    \vspace{-4mm}
    \caption{Nine linearly sampled frames of our method. Painting order from left to right, top to bottom. The image with the green border is the reference painting.}
    \label{fig:appx:ours_1}
\end{figure*}

\begin{figure*}[htbp!]
    \centering

    \begin{subfigure}[b]{\textwidth}
        \centering
        \includegraphics[width=\textwidth]{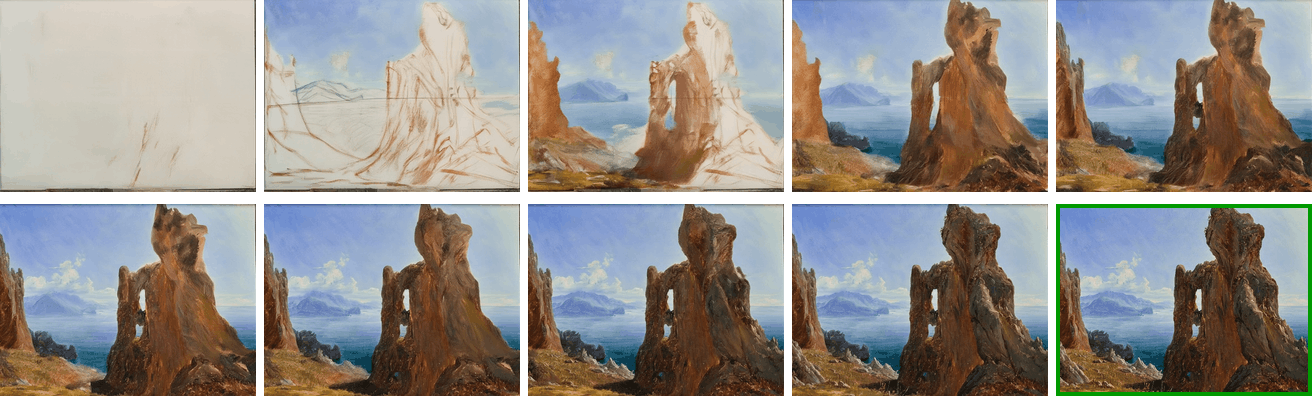}
        \vspace{-5mm}
        \caption{\textit{Arco Naturale, Capri} by Thomas Fearnley, Public Domain, via Wikimedia Commons.}
    \end{subfigure}
    
    \vspace{2mm}
    \begin{subfigure}[b]{\textwidth}
        \centering
        \includegraphics[width=\textwidth]{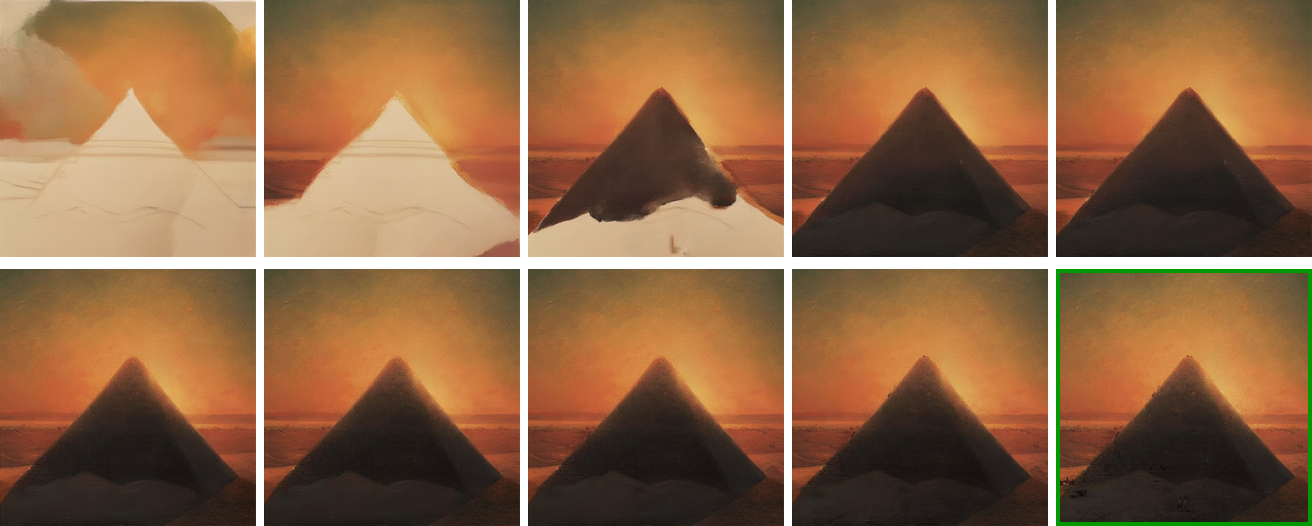}
        \vspace{-5mm}
        \caption{\textit{The Great Pyramid at Giza} by Ivan Aivazovsky, Public Domain, via Wikimedia Commons.}
    \end{subfigure}

    \vspace{2mm}
    \begin{subfigure}[b]{\textwidth}
        \centering
        \includegraphics[width=\textwidth]{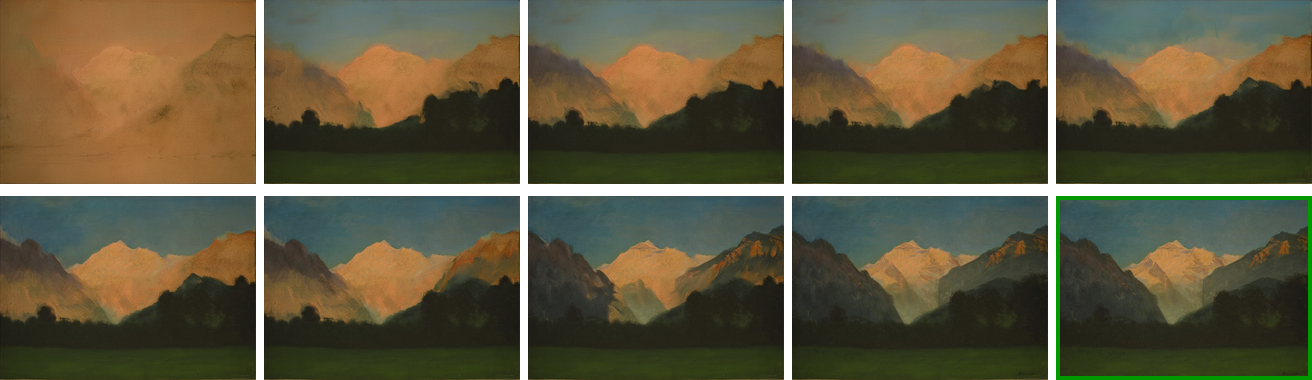}
        \vspace{-5mm}
        \caption{\textit{View of Glacier Park or Sunset on Peak} by Albert Bierstadt, Public Domain, via Wikimedia Commons.}
    \end{subfigure}

    \vspace{-1mm}
    \caption{Nine linearly sampled frames of our method. Painting order from left to right, top to bottom. The image with the green border is the reference painting.}
    \label{fig:appx:ours_2}
\end{figure*}

\begin{figure*}[htbp!]
    \centering

    \begin{subfigure}[b]{\textwidth}
        \centering
        \includegraphics[width=0.91\textwidth]{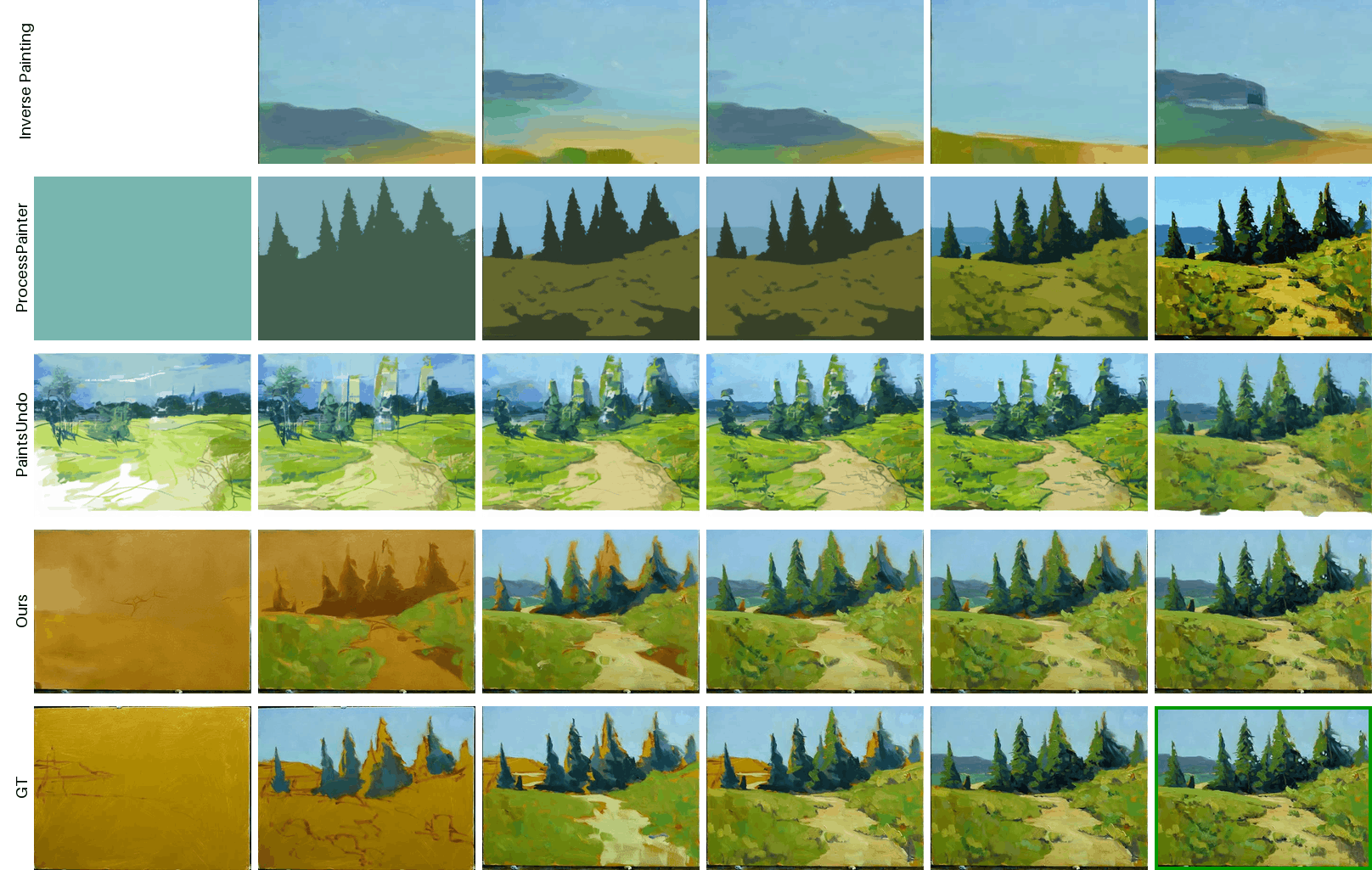}
    \end{subfigure}
    
    \vspace{8mm}
    \begin{subfigure}[b]{\textwidth}
        \centering
        \includegraphics[width=0.91\textwidth]{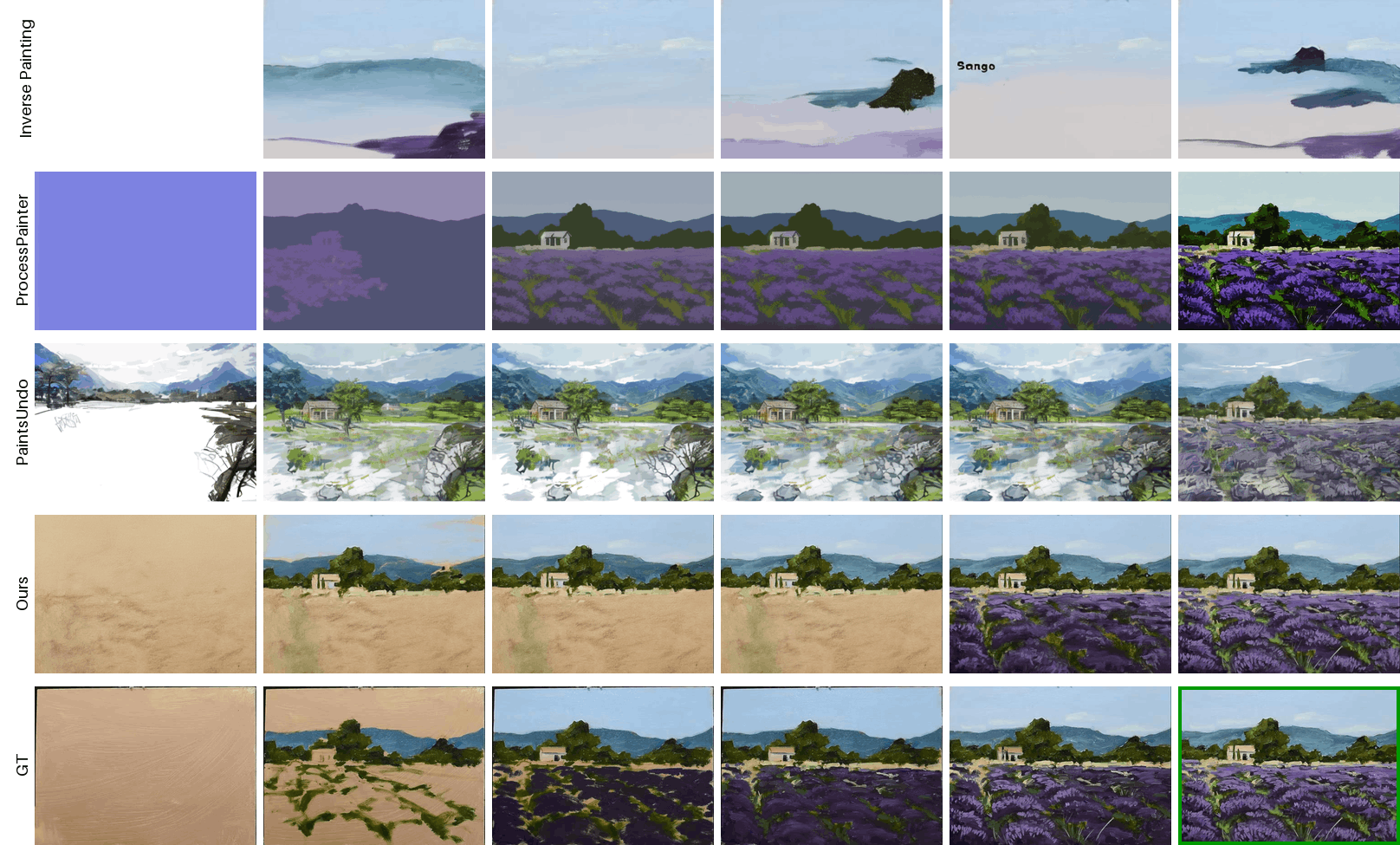}
    \end{subfigure}

    \vspace{-1mm}
    \caption{Comparison with ground truth (GT) of our collected dataset. Painting order from left to right, GT frame with green border is the reference frame. Image courtesy of Samir Godinjak, from the Painting with Samir YouTube channel}
    \label{fig:appx:comparison_gt}
\end{figure*}

\begin{figure*}[htbp!]
    \centering

    \begin{subfigure}[b]{\textwidth}
        \centering
        \includegraphics[width=\textwidth]{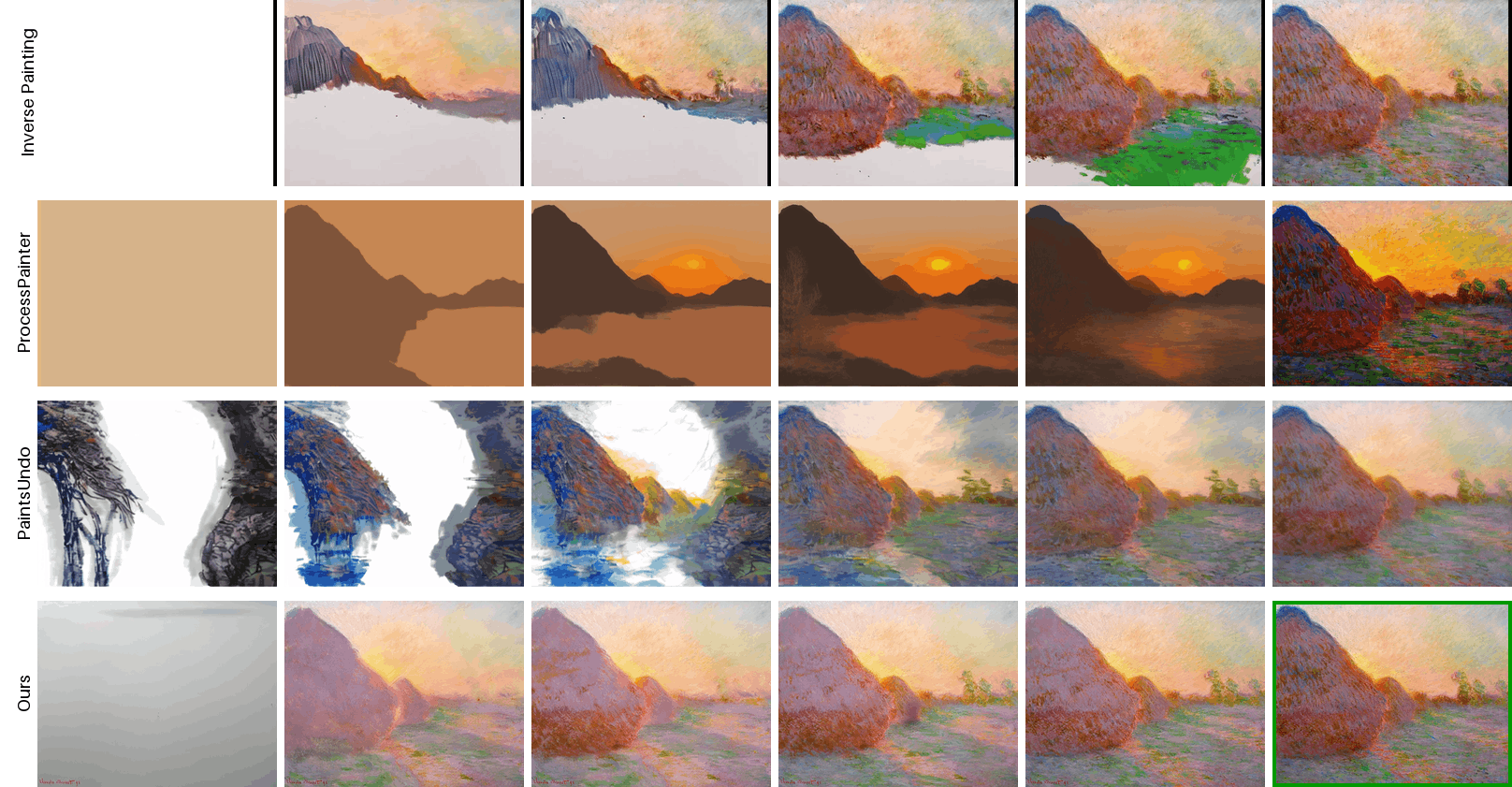}
        \vspace{-4mm}
        \caption{Human evaluation win rates: Inverse Painting (53\%), ProcessPainter (13\%), Paints-Undo (13\%), and Ours (100\%). \textit{Meules} by Claude Monet, Public Domain, via Wikimedia Commons.}
    \end{subfigure}
    
    \vspace{5mm}
    \begin{subfigure}[b]{\textwidth}
        \centering
        \includegraphics[width=\textwidth]{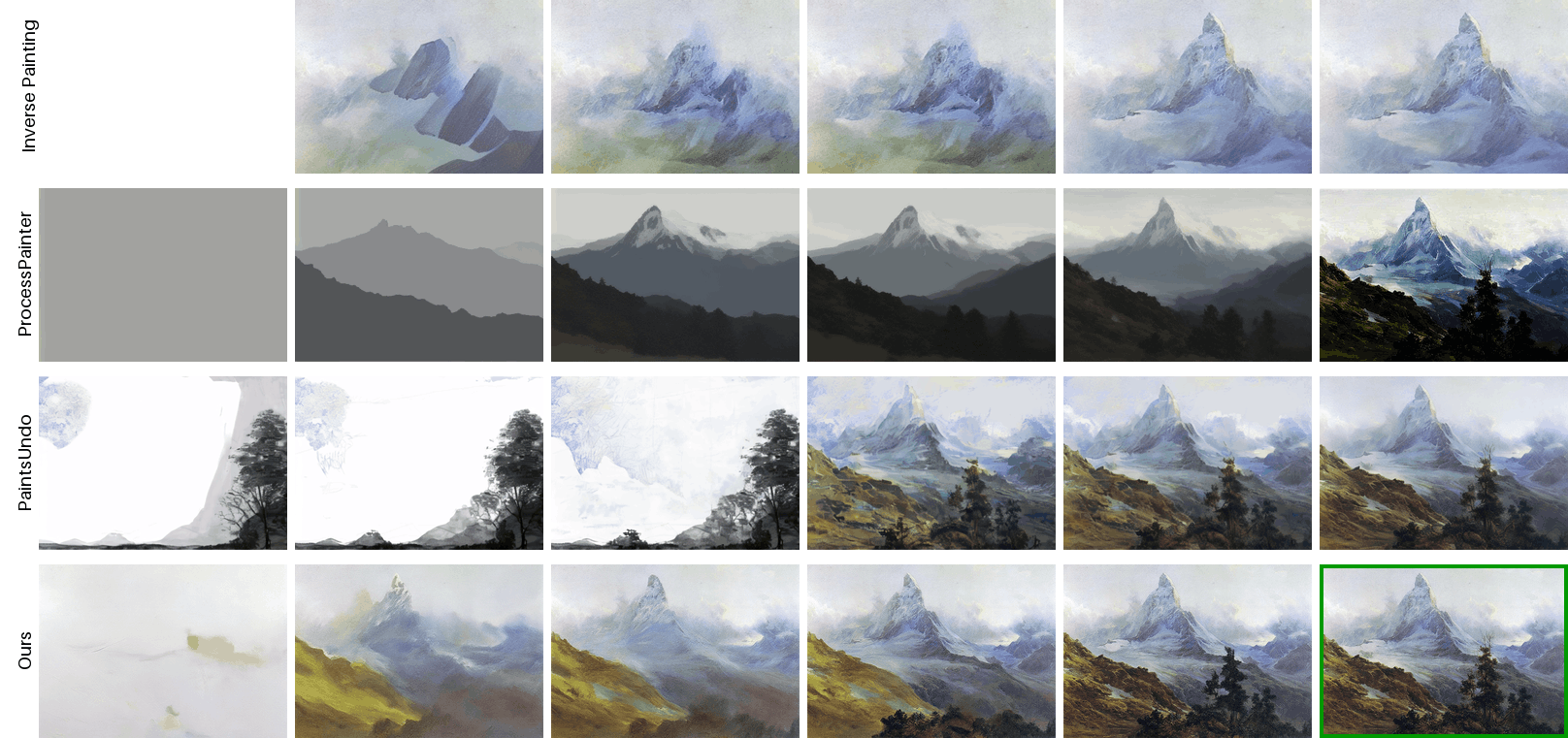}
        \vspace{-4mm}
        \caption{Human evaluation win rates: Inverse Painting (16\%), ProcessPainter (39\%), Paints-Undo (27\%), and Ours (95\%). \textit{Matterhorn} by Edward Theodore Compton, Public Domain, via Wikimedia Commons.}
    \end{subfigure}

    \caption{Method comparison. Painting order from left to right. The final frame (green border) is visually nearly indistinguishable from the reference image; the latter is omitted for brevity.}
    \label{fig:appx:comparison_1}
\end{figure*}

\begin{figure*}[htbp!]
    \centering

    \begin{subfigure}[b]{\textwidth}
        \centering
        \includegraphics[width=0.8\textwidth]{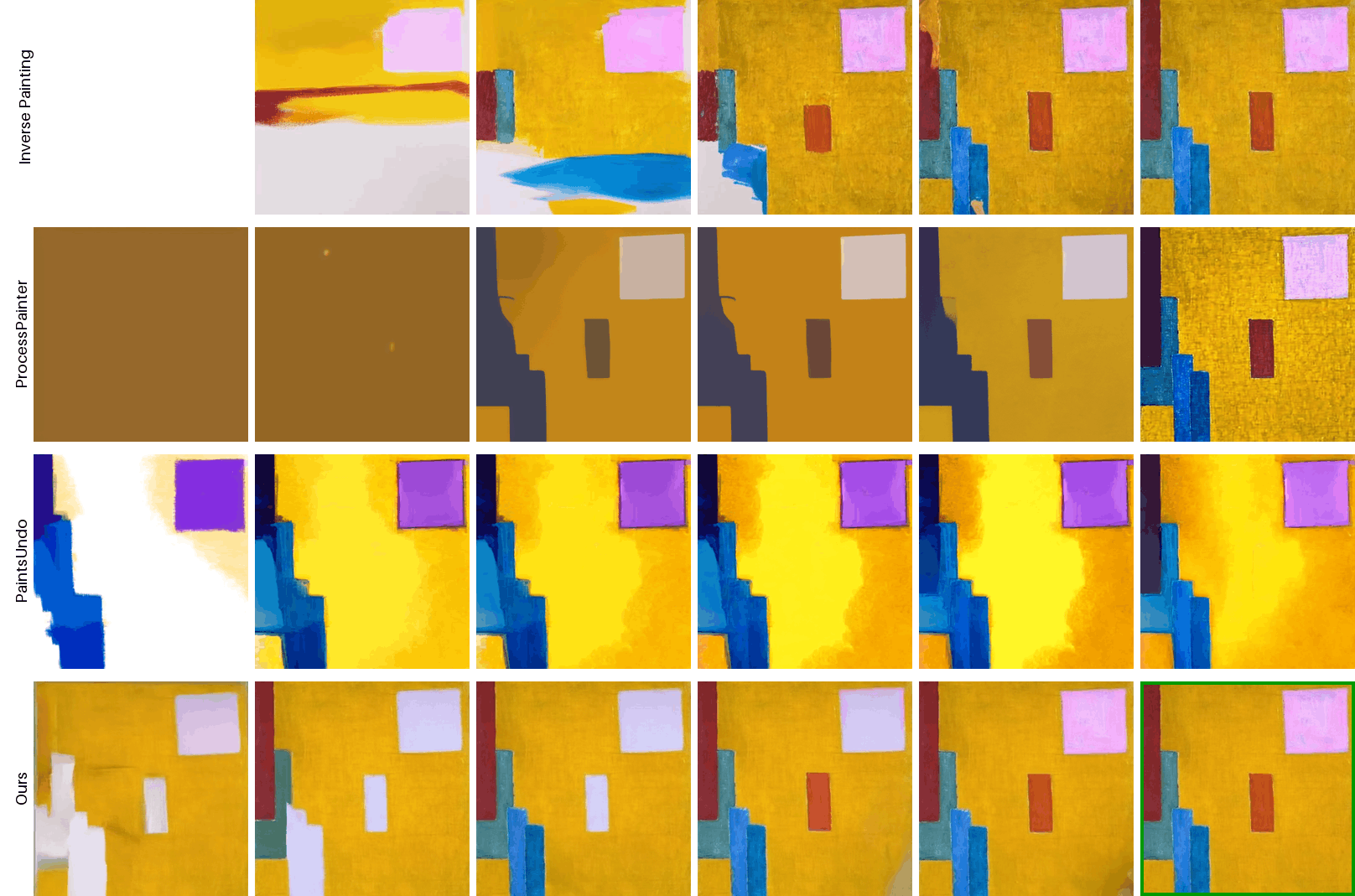}
        \caption{Our method shows the best understanding of the geometric objects. Human evaluation win rates favor our approach: Inverse Painting (31.6\%), ProcessPainter (38.1\%), Paints-Undo (16.7\%), and Ours (100\%). \textit{Abstract Painting} by Vanessa Bell, Public Domain, via Wikiart.}
    \end{subfigure}
    
    \vspace{3mm}
    \begin{subfigure}[b]{\textwidth}
        \centering
        \includegraphics[width=\textwidth]{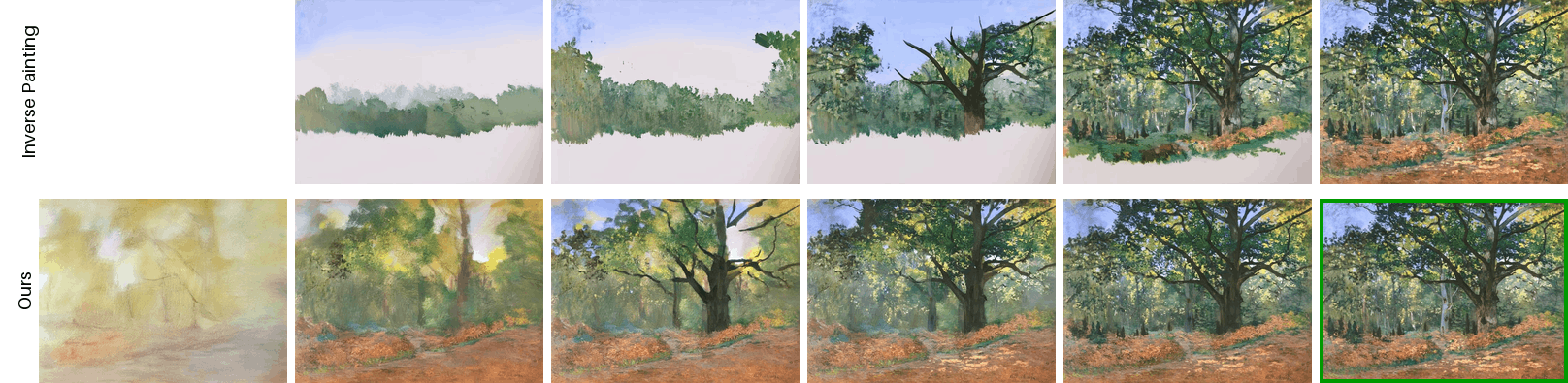}
        \vspace{-5mm}
        \caption{Win rates: Inverse Painting (76\%) and Ours (75\%). \textit{The Bodmer Oak} by Claude Monet, Public Domain, via Wikimedia Commons.}
    \end{subfigure}

    \vspace{3mm}
    \begin{subfigure}[b]{\textwidth}
        \centering
        \includegraphics[width=\textwidth]{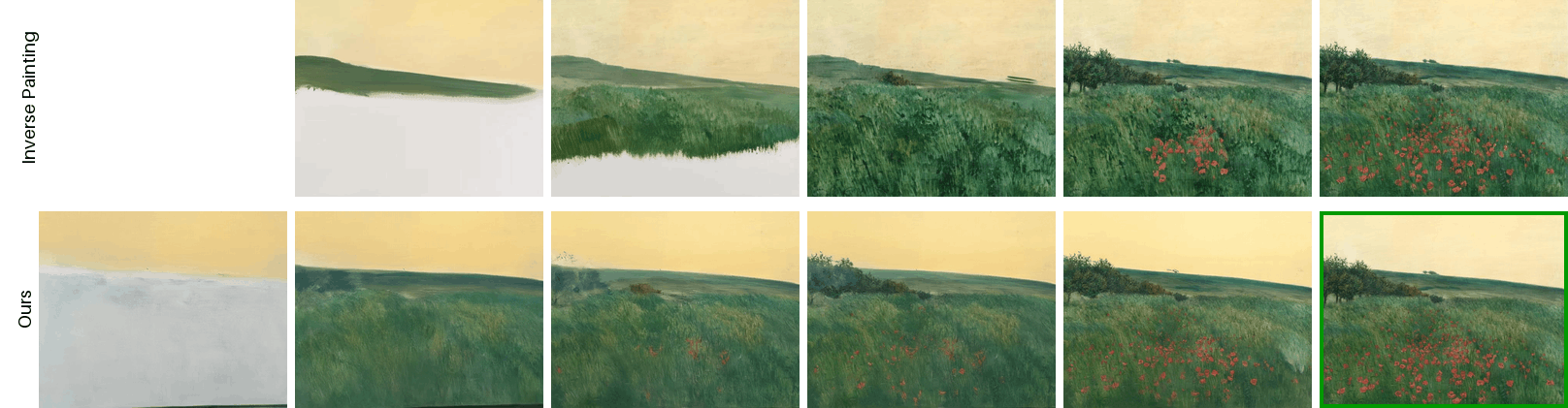}
        \vspace{-5mm}
        \caption{Win rates: Inverse Painting (61\%) and Ours (65\%). \textit{Red Poppies} by František Kaván, Public Domain, via Wikimedia Commons.}
    \end{subfigure}

    \vspace{-3mm}
    \caption{Method comparison. Painting order from left to right. The final frame (green border) is visually nearly indistinguishable from the reference image; the latter is omitted for brevity.}
    \label{fig:appx:comparison_2}
\end{figure*}

\begin{figure*}[htbp!]
    \centering

    \begin{subfigure}[b]{0.95\textwidth}
        \centering
        \includegraphics[width=\textwidth]{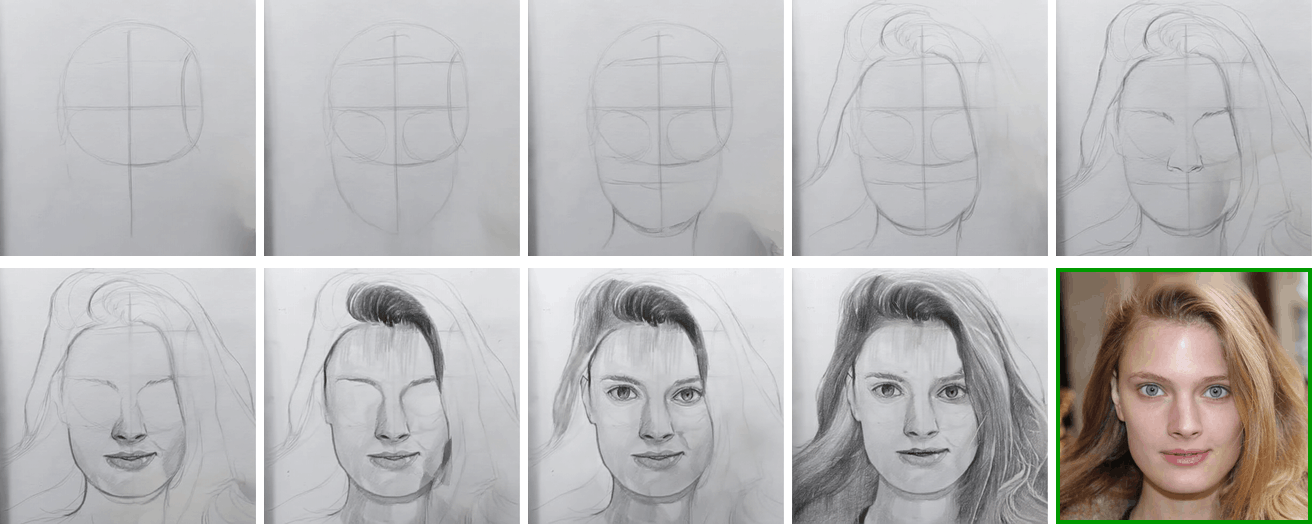}
    \end{subfigure}
    
    \vspace{3mm}
    \begin{subfigure}[b]{0.95\textwidth}
        \centering
        \includegraphics[width=\textwidth]{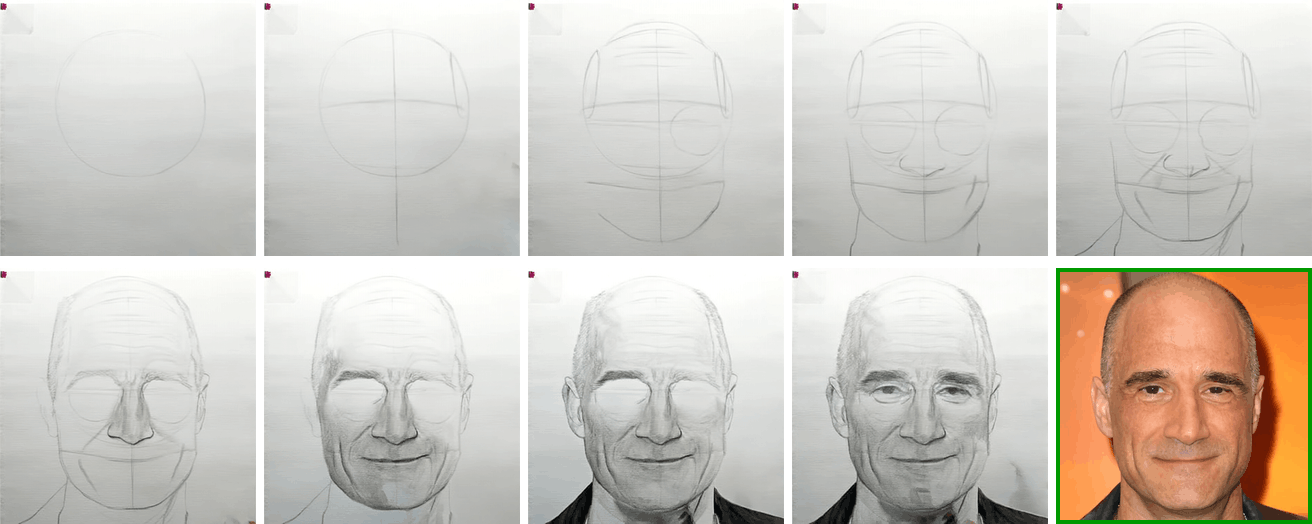}
    \end{subfigure}

    \vspace{3mm}
    \begin{subfigure}[b]{0.95\textwidth}
        \centering
        \includegraphics[width=\textwidth]{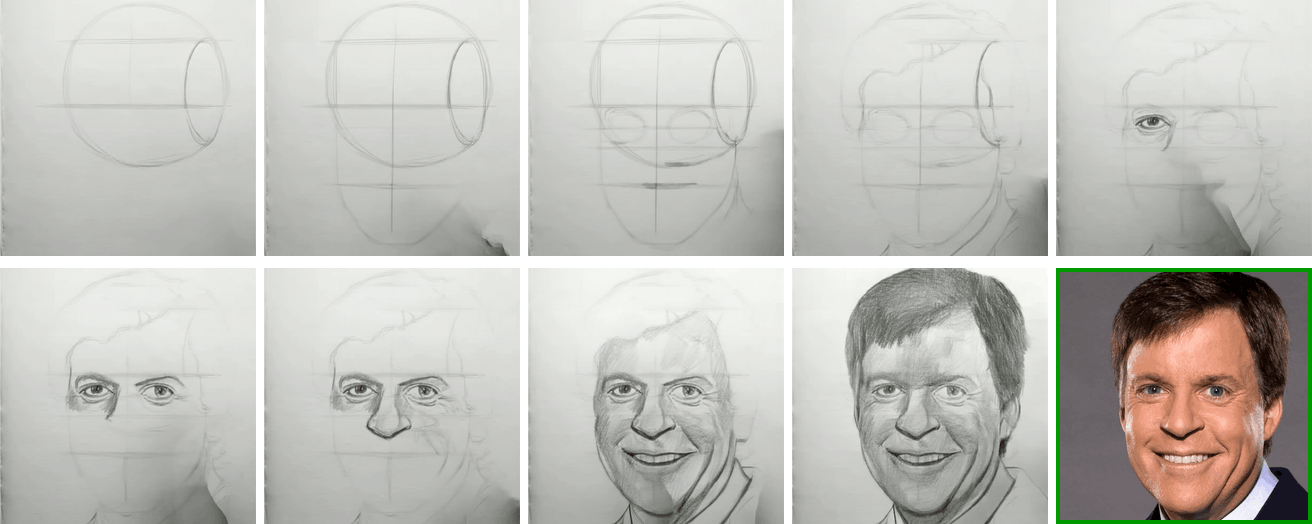}
    \end{subfigure}

    \vspace{-3mm}
    \caption{Loomis portrait drawings on CelebA-HQ \cite{karras2017progressive}. The sampled keyframes illustrate the progressive, step-by-step construction of the faces using the Loomis method. The reference image is bordered in green.}
    \label{fig:appx:loomis_celeba_hq}
\end{figure*}

\begin{figure*}[htbp!]
    \centering

    \begin{subfigure}[b]{0.95\textwidth}
        \centering
        \includegraphics[width=\textwidth]{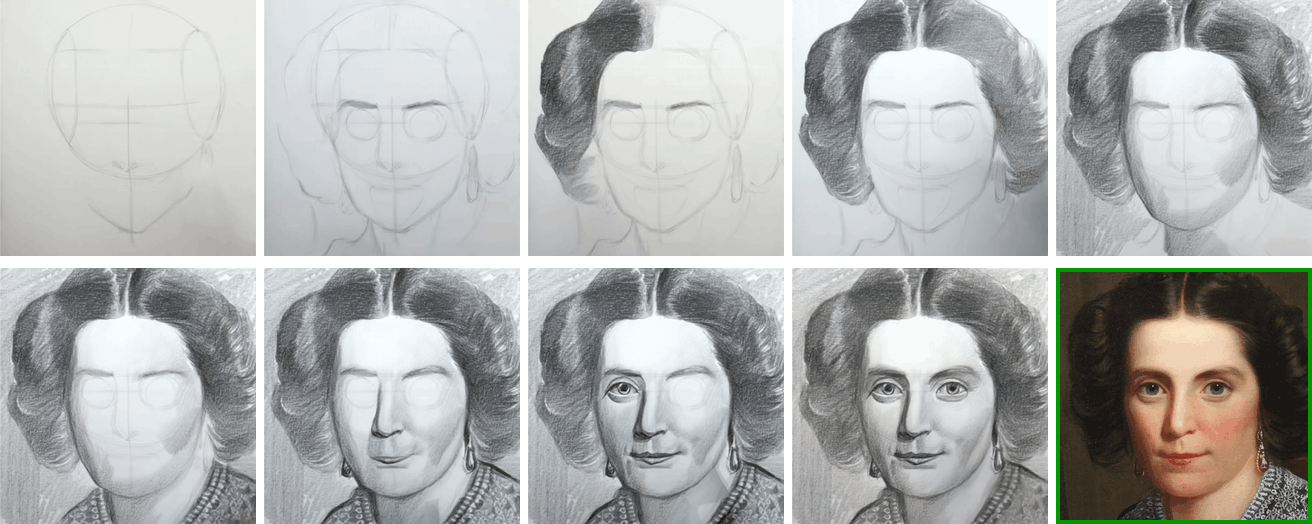}
    \end{subfigure}
    
    \vspace{3mm}
    \begin{subfigure}[b]{0.95\textwidth}
        \centering
        \includegraphics[width=\textwidth]{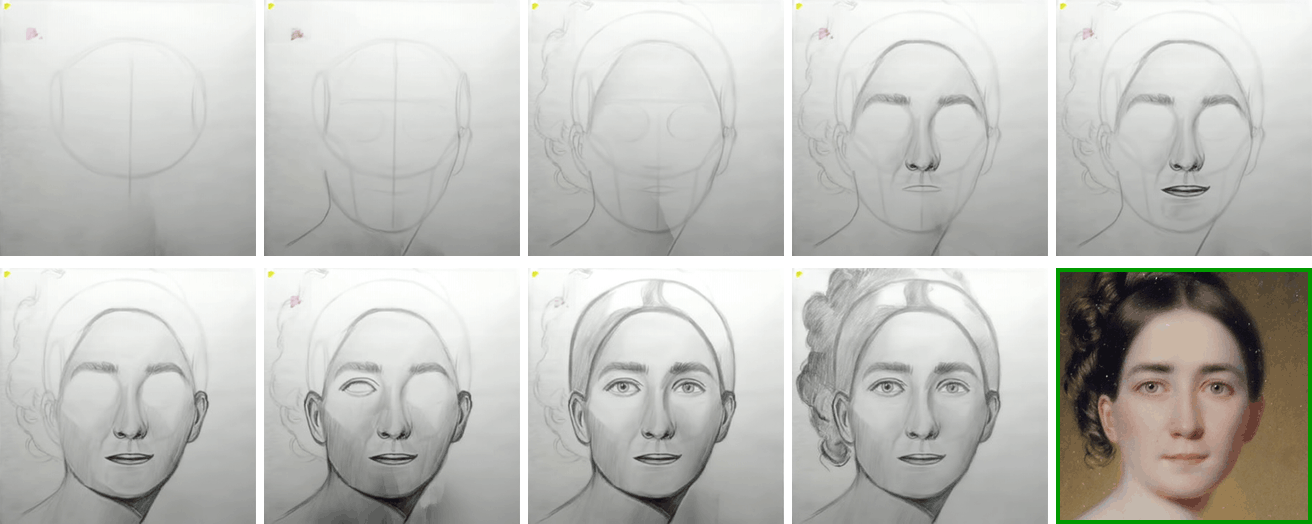}
    \end{subfigure}

    \vspace{3mm}
    \begin{subfigure}[b]{0.95\textwidth}
        \centering
        \includegraphics[width=\textwidth]{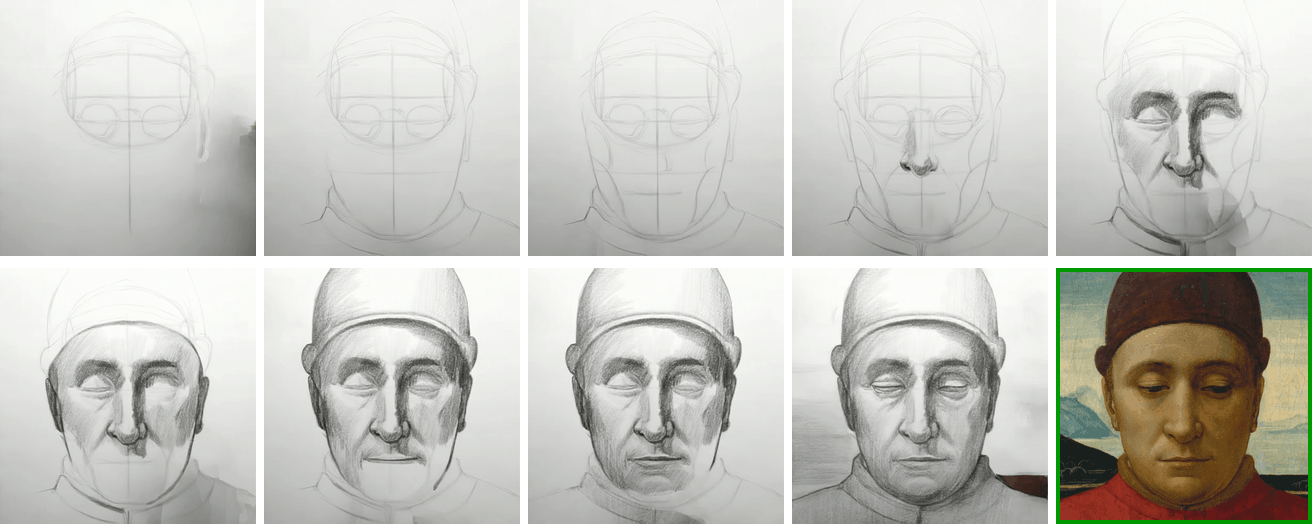}
    \end{subfigure}

    \vspace{-3mm}
    \caption{Loomis portrait drawings on MetFaces \cite{karras2020training}. The sampled keyframes illustrate the progressive, step-by-step construction of the faces using the Loomis method. The reference image is bordered in green.}
    \label{fig:appx:loomis_art_met}
\end{figure*}

\begin{figure*}[htbp!]
    \centering

    \begin{subfigure}[b]{0.95\textwidth}
        \centering
        \includegraphics[width=\textwidth]{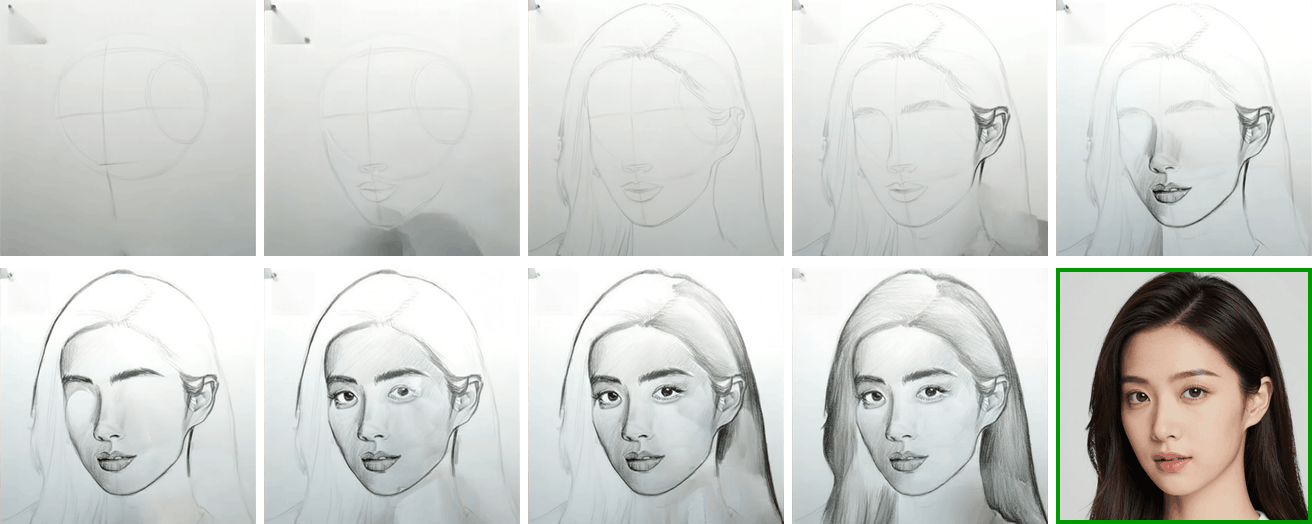}
    \end{subfigure}
    
    \vspace{3mm}
    \begin{subfigure}[b]{0.95\textwidth}
        \centering
        \includegraphics[width=\textwidth]{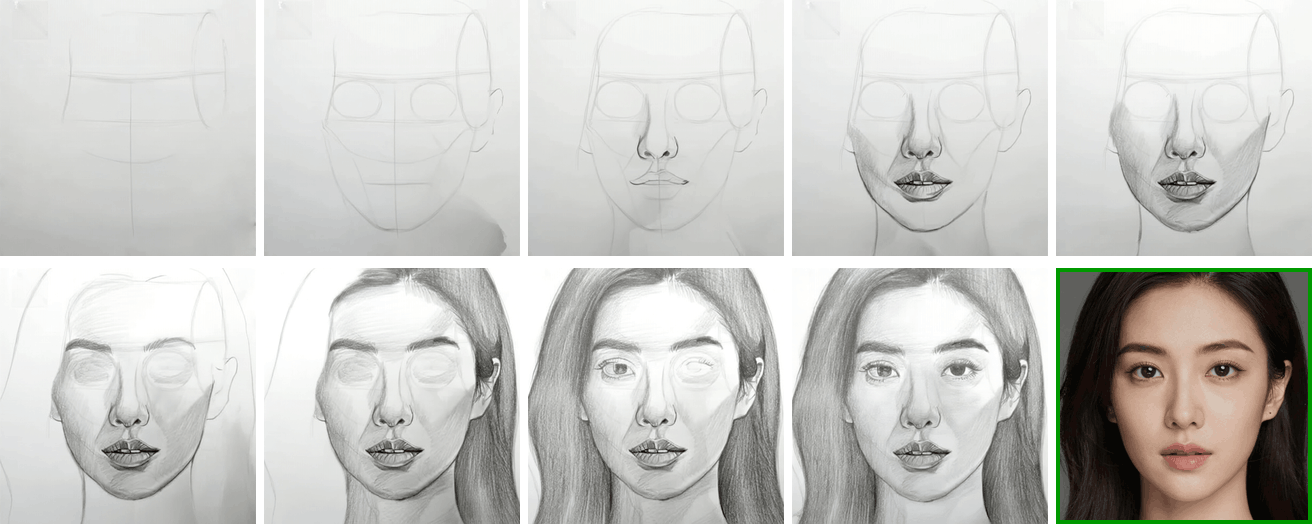}
    \end{subfigure}

    \vspace{3mm}
    \begin{subfigure}[b]{0.95\textwidth}
        \centering
        \includegraphics[width=\textwidth]{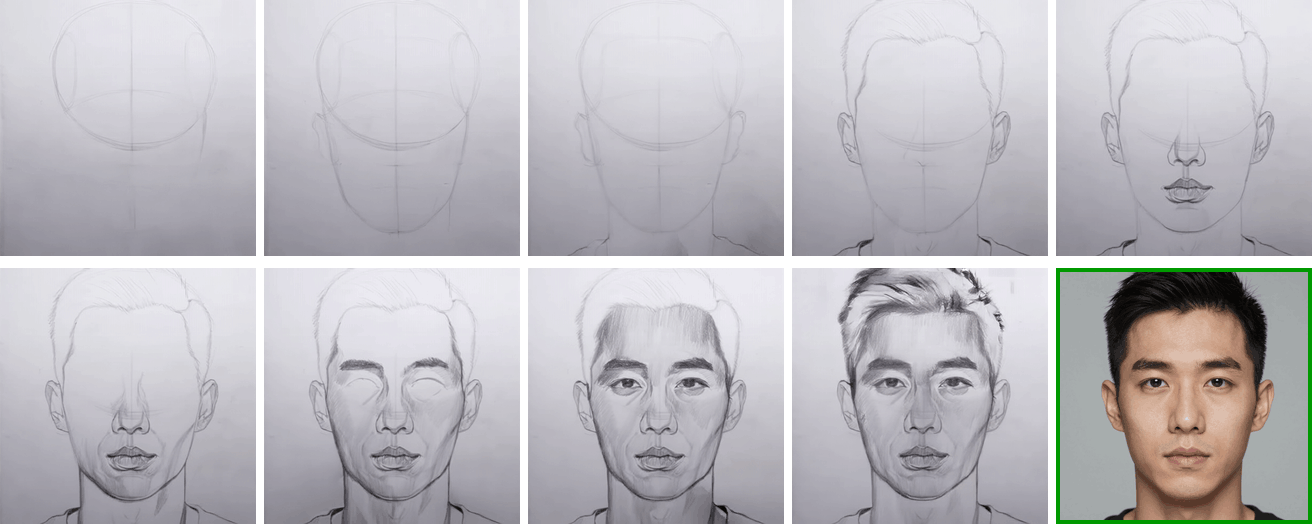}
    \end{subfigure}

    \vspace{-3mm}
    \caption{Loomis portraits. The sampled keyframes illustrate the progressive, step-by-step construction of the faces using the Loomis method. The reference image, generated with Z Image Turbo \cite{cai2025z}, bordered in green.}
    \label{fig:appx:loomis_z_image}
\end{figure*}

\begin{figure*}[htbp!]
    \centering

    \begin{subfigure}[b]{\textwidth}
        \centering
        \includegraphics[width=\textwidth]{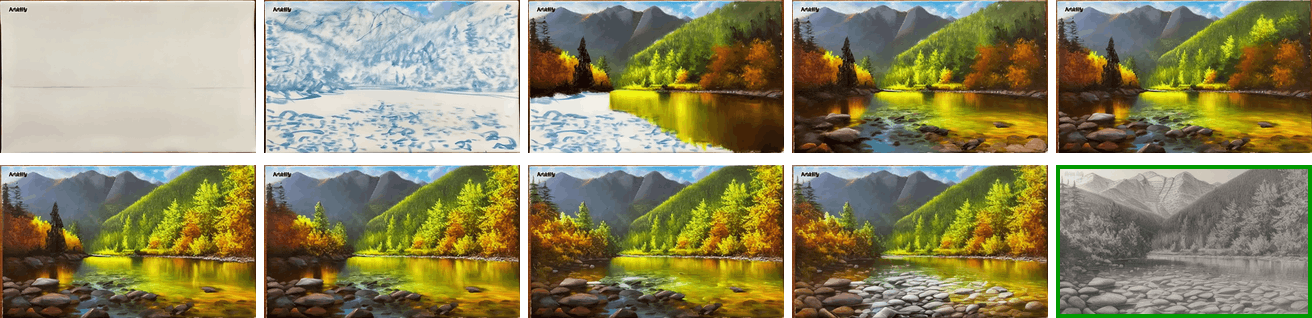}
        \vspace{-5mm}
        \caption{Art media transfer, pencil sketch to acrylic painting. Input image is a style-transferred version from \cite{aluart_stream} via ControlNet.}
    \end{subfigure}
    
    \vspace{2mm}
    \begin{subfigure}[b]{\textwidth}
        \centering
        \includegraphics[width=\textwidth]{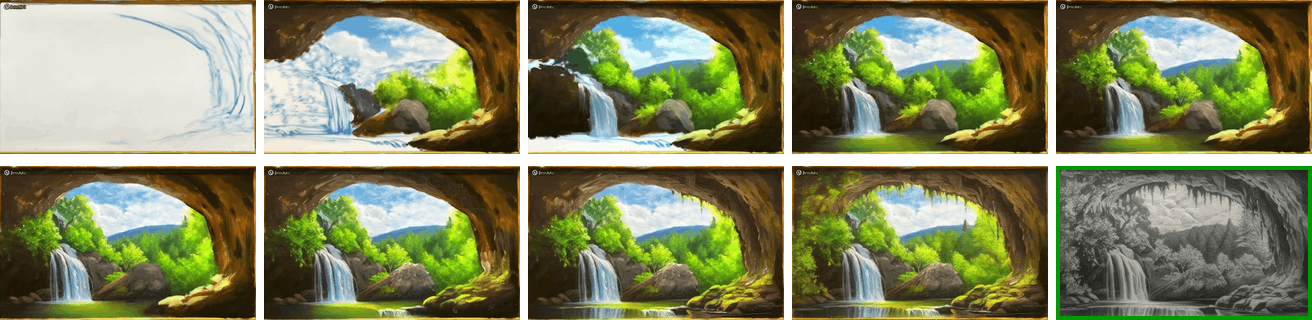}
        \vspace{-5mm}
        \caption{Art media transfer, pencil sketch to acrylic painting. Input image is a style-transferred version from \cite{aluart_cave} via ControlNet.}
    \end{subfigure}

    \vspace{2mm}
    \begin{subfigure}[b]{\textwidth}
        \centering
        \includegraphics[width=\textwidth]{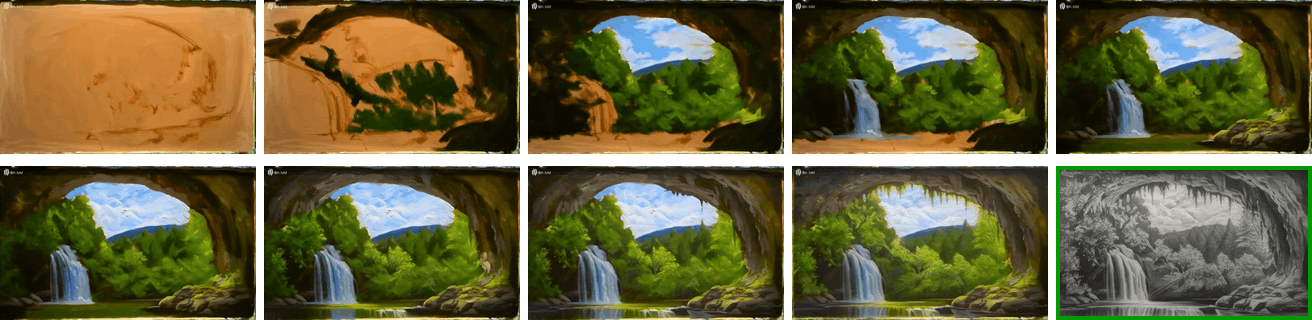}
        \vspace{-5mm}
        \caption{Art media transfer, pencil sketch to oil painting. Input image is a style-transferred version from \cite{aluart_cave} via ControlNet.}
    \end{subfigure}

    \vspace{2mm}
    \begin{subfigure}[b]{\textwidth}
        \centering
        \includegraphics[width=\textwidth]{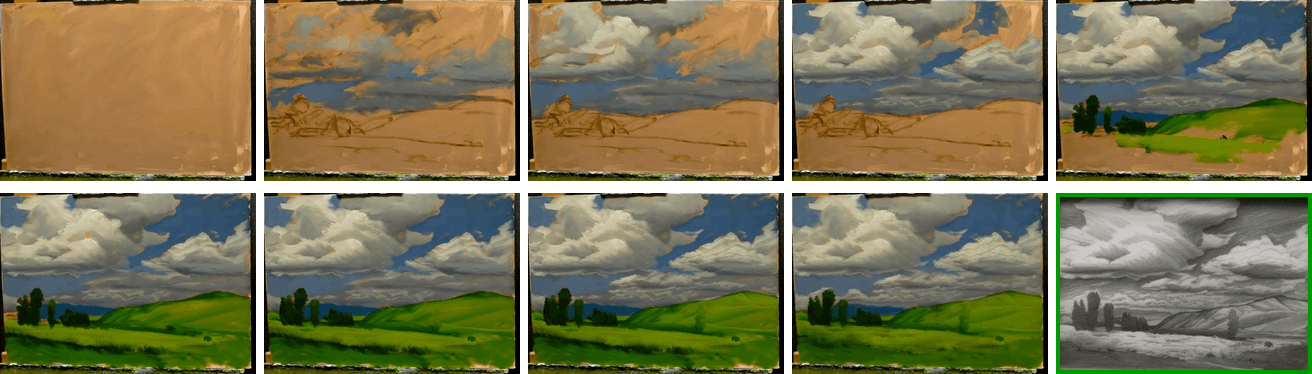}
        \vspace{-5mm}
        \caption{Art media transfer, pencil sketch to oil painting. Input image is a style-transferred version from \cite{samir2024small} via ControlNet.}
    \end{subfigure}

    \vspace{-1mm}
    \caption{Art media transfer, from pencil to acrylic and oil. Nine linearly sampled frames of our method. Painting order from left to right, top to bottom. The image with the green border is the reference painting.}
    \label{fig:appx:art_media_transfer_ao}
\end{figure*}

\begin{figure*}[htbp!]
    \centering

    \begin{subfigure}[b]{\textwidth}
        \centering
        \includegraphics[width=\textwidth]{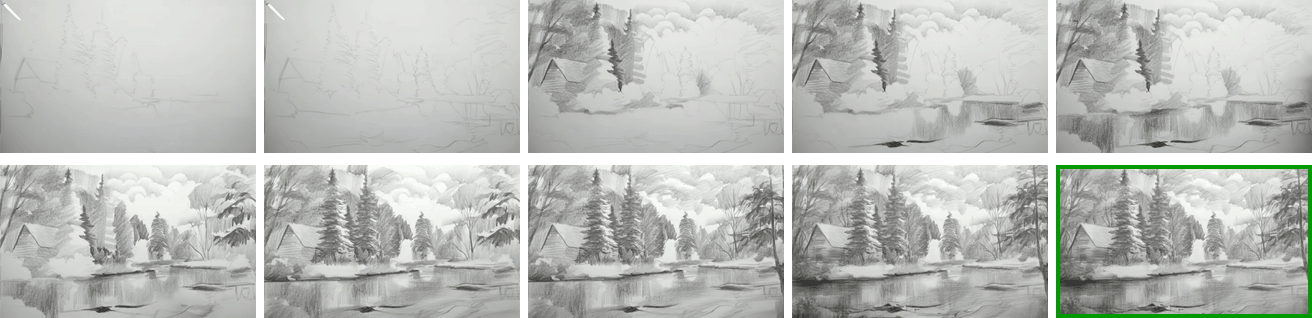}
        \vspace{-5mm}
        \caption{Reconstruction of the Painting process. Input image is a style-transferred version from \cite{aluart_winter} via ControlNet.}
    \end{subfigure}
    
    \vspace{2mm}
    \begin{subfigure}[b]{\textwidth}
        \centering
        \includegraphics[width=\textwidth]{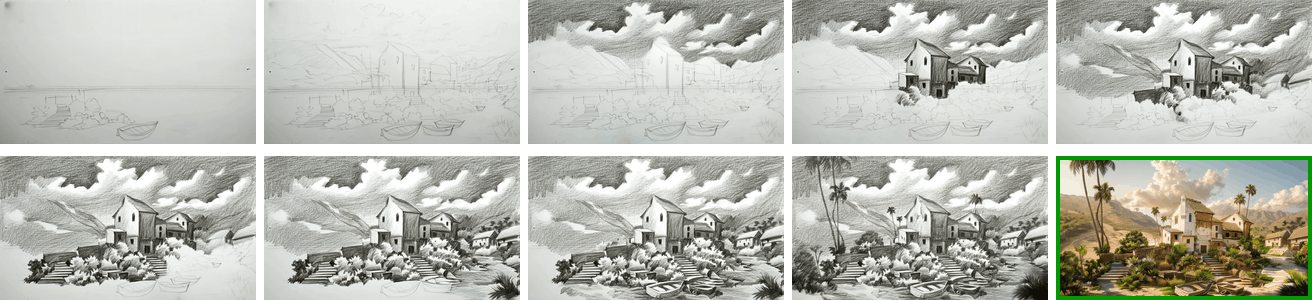}
        \vspace{-5mm}
        \caption{Art media transfer, color painting to pencil sketch. Input image is a style-transferred version from \cite{paintlane2017village} via ControlNet.}
    \end{subfigure}

    \vspace{2mm}
    \begin{subfigure}[b]{\textwidth}
        \centering
        \includegraphics[width=\textwidth]{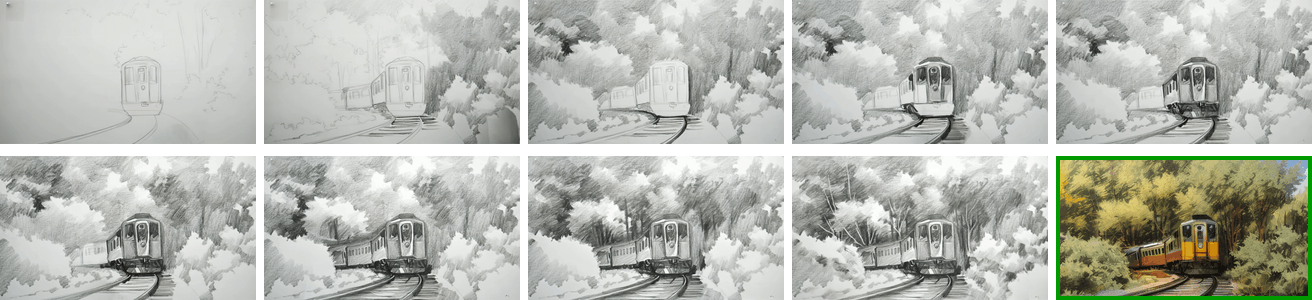}
        \vspace{-5mm}
        \caption{Art media transfer, color painting to pencil sketch. Input image is a style-transferred version from \cite{paintlane2022train} via ControlNet.}
    \end{subfigure}

    \vspace{-1mm}
    \caption{Pencil paintings. Nine linearly sampled frames of our method. Painting order from left to right, top to bottom. The image with the green border is the reference painting.}
    \label{fig:appx:art_media_transfer_p}
\end{figure*}

We present qualitative results across a diverse set of input images. First, we showcase our model's reconstruction capabilities on a wide range of landscape paintings in \cref{fig:appx:ours_1} and \cref{fig:appx:ours_2}. Next, visual comparisons against the evaluated baselines are provided in \cref{fig:appx:comparison_gt}, \cref{fig:appx:comparison_1}, and \cref{fig:appx:comparison_2}. Furthermore, \cref{fig:appx:loomis_celeba_hq}, \cref{fig:appx:loomis_art_met}, and \cref{fig:appx:loomis_z_image} demonstrate the generation of pencil sketches using the Loomis portrait method, evaluating this approach on real photographs, paintings, and synthesized images. Finally, we illustrate the effectiveness of our art media transfer model in \cref{fig:appx:art_media_transfer_ao} and \cref{fig:appx:art_media_transfer_p}, highlighting its ability to seamlessly bridge different artistic media during the generation process.

\begin{figure}
    \centering
    \begin{subfigure}[b]{\linewidth}
        \centering
        \includegraphics[width=\linewidth]{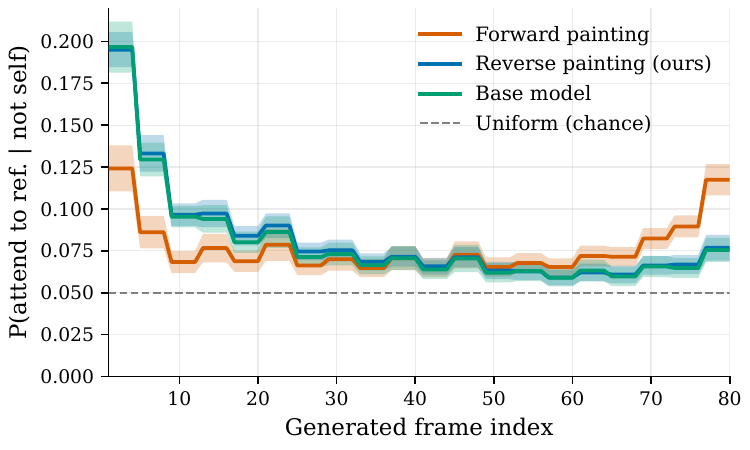}
        \caption{Attention probabilities from the Vace 1.3B model not considering attending to the current frame to generate.}
    \end{subfigure}
    
    \begin{subfigure}[b]{\linewidth}
        \centering
        \includegraphics[width=\linewidth]{figures/attn_vis_figures/reference_decay_all_three_5B.pdf}
        \caption{Attention probabilities from the Wan 2.2 I2V 5B model not considering attending to the current frame to generate.}
    \end{subfigure}

    \begin{subfigure}[b]{\linewidth}
        \centering
        \includegraphics[width=\linewidth]{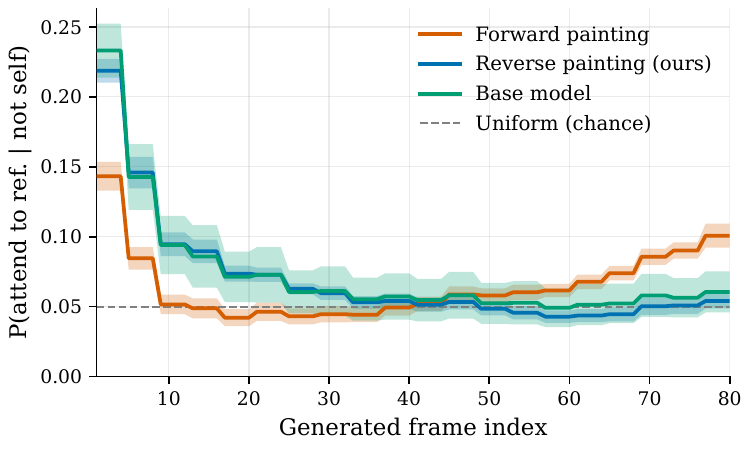}
        \caption{Attention probabilities from the Wan 2.1 I2V 14B model not considering attending to the current frame to generate.}
    \end{subfigure}
    \vspace{-5mm}
    \caption{Reference attention distribution. Probability of attending to the reference frame relative to non-self frames ($P(\text{attend to ref.} \mid \text{not self})$). Our reverse painting model maintains a decaying attention profile similar to the base model across architectures, whereas forward painting shifts focus heavily toward later frames. Fine-tuned models were trained for 7 epochs. Stepwise plateaus of 4 frames reflect the model's $4\times$ temporal latent compression.}
    \label{fig:appx:attn_vis_figure}
    \vspace{-6mm}
\end{figure}

\begin{table*}[htpb!]
    \centering
    \setlength{\tabcolsep}{5pt}
    \begin{tabular}{l|ccccc|ccc|r}
        \toprule
        \textbf{Method} & \textbf{FID $\downarrow$} & \textbf{LPIPS $\downarrow$} & \textbf{CLIP $\uparrow$} & \textbf{DINOv2 $\uparrow$} & \textbf{DDC $\downarrow$} & \multicolumn{3}{c|}{\textbf{Distance}} & \textbf{Params} \\
        & & & & & & \textbf{LPIPS $\downarrow$} & \textbf{CLIP $\uparrow$} & \textbf{DINOv2 $\uparrow$} & \\
        \midrule
        Qwen + Base        & 306.03 & 0.70 & 0.74 & 0.52 & 41.16 & 0.68 & 0.86 & 0.79 & 20B + 5B \\
        Fine tuned + Base  & 169.32 & 0.45 & \textbf{0.84} & \textbf{0.73} & 11.97 & 0.24 & 0.96 & \textbf{0.97} &  5B + 5B \\
        Art Media Transfer & \textbf{168.78} & \textbf{0.44} & \textbf{0.84} & \textbf{0.73} & \textbf{10.36} & \textbf{0.22} & \textbf{0.97} & \textbf{0.97} &  5B      \\
        \midrule
        Oracle + Base      & 152.42 & 0.38 & 0.85 & 0.75 &  7.97 & 0.03 & 0.98 & 0.99 &  5B      \\
        \bottomrule
    \end{tabular}
    \caption{Quantitative comparison of our proposed Art Media Transfer model against baselines. Best non-oracle results are in \textbf{bold}. Art Media Transfer achieves the highest overall performance while requiring significantly fewer parameters. ``Distance'' measures visual similarity between the generated final frame and the ground truth. Oracle + Base establishes the theoretical upper bound.}
    \label{tab:appx:art_media_transfer}
\end{table*}

\subsection{Art Media Transfer}
\label{sec:appx:art_media_transfer}
To evaluate our art media transfer framework, we test whether a model can revert a style-transferred image back to its target art medium while simultaneously reconstructing its step-by-step painting process. We compare our approach against two-stage baselines, defined as follows:

\begin{itemize}[leftmargin=*]
    \item \textbf{Qwen Image Edit + Base}: A two-stage, few-shot baseline using Qwen Image Edit~\cite{wu2025qwen}. Given a style-transferred image and an exemplar from the target medium, Qwen is prompted to transfer the style of the exemplar onto the input image. In the second stage, our base model predicts the sequence from the edited result.
    \item \textbf{Fine-Tuned I2I + Base}: A dedicated, purpose-built single-frame image-to-image (I2I) baseline. To ensure a fair comparison, we fine-tune a separate Wan 2.2 5B I2V model to output a single image instead of a video sequence, using the exact same hyperparameters, training epochs, and prompts as our media transfer model. The fine-tuned I2I model first converts the image to the target medium, followed by our base model generating the full painting process.
\end{itemize}

\noindent We evaluate these baselines against our unified \textbf{Art Media Transfer} model, which performs art media conversion and painting process reconstruction in a single pass using Wan 2.2 5B I2V as the backbone. To establish a theoretical upper bound, we also report the performance of our base model evaluated directly on the ground-truth reference images (\textbf{Oracle + Base}).

\noindent In addition to the introduced visual quality metrics, we evaluate the visual Distance (LPIPS, CLIP, and DINOv2) between the generated final painting frame and the ground-truth target image to measure media translation fidelity.

\begin{figure*}[htbp!]
    \centering

    \begin{subfigure}[b]{\textwidth}
        \centering
        \includegraphics[width=\textwidth]{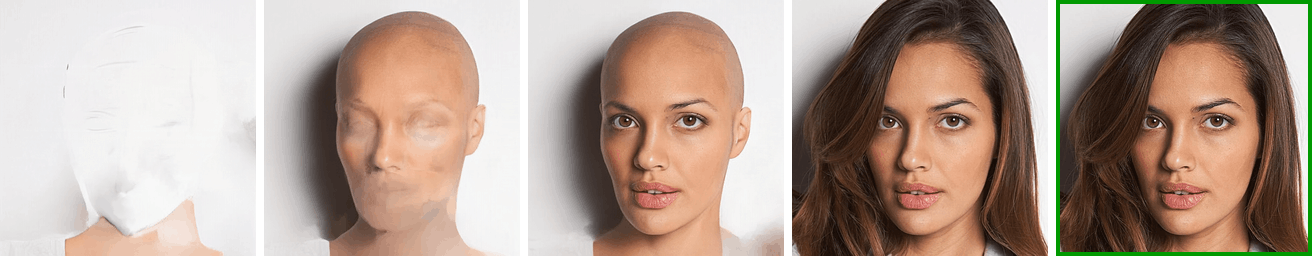}
        \vspace{-4mm}
        \caption{  
            Base model generation on an out-of-domain portrait photograph. Having not been exposed to real photographs during fine-tuning, the model struggles to synthesize a painting process. However, it still exhibits a fundamental understanding of rendering order, reserving fine details like facial features and hair for the final stages.
        }
    \end{subfigure}
    
    \vspace{3mm}
    \begin{subfigure}[b]{\textwidth}
        \centering
        \includegraphics[width=\textwidth]{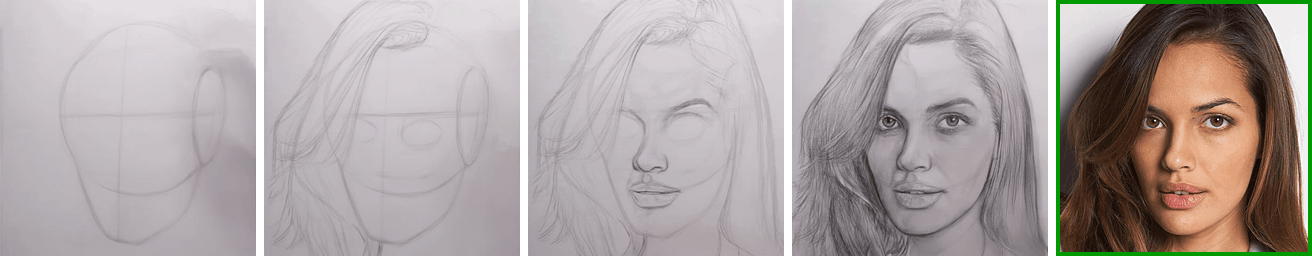}
        \vspace{-4mm}
        \caption{
            Art media transfer model generation on the same photograph. The model successfully employs structural guidelines, demonstrating its ability to generate pencil sketches from photographic inputs.}
    \end{subfigure}

    \caption{
    Qualitative comparison demonstrating the effectiveness of the proposed art media transfer framework. While the base model reconstruct the input, the art media transfer model enables the application of structured artistic procedures, such as the Loomis portrait method, to real-world photographic inputs. Reference photo from CelebA-HQ \cite{karras2017progressive}.}
    \label{fig:appx:base_vs_amt}
\end{figure*}

\noindent As shown in \cref{tab:appx:art_media_transfer}, our unified Art Media Transfer model matches or exceeds the Fine-Tuned I2I + Base baseline across all evaluation metrics, coming closest to the Oracle upper bound, despite using half the parameters and a single model rather than a dedicated conversion stage. This indicates that a single model can jointly learn medium conversion and process reconstruction without sacrificing quality relative to a specialized two-stage pipeline. A visual comparison against the base model is shown in \cref{fig:appx:base_vs_amt}.

\begin{figure}
    \centering
    \includegraphics[width=\linewidth]{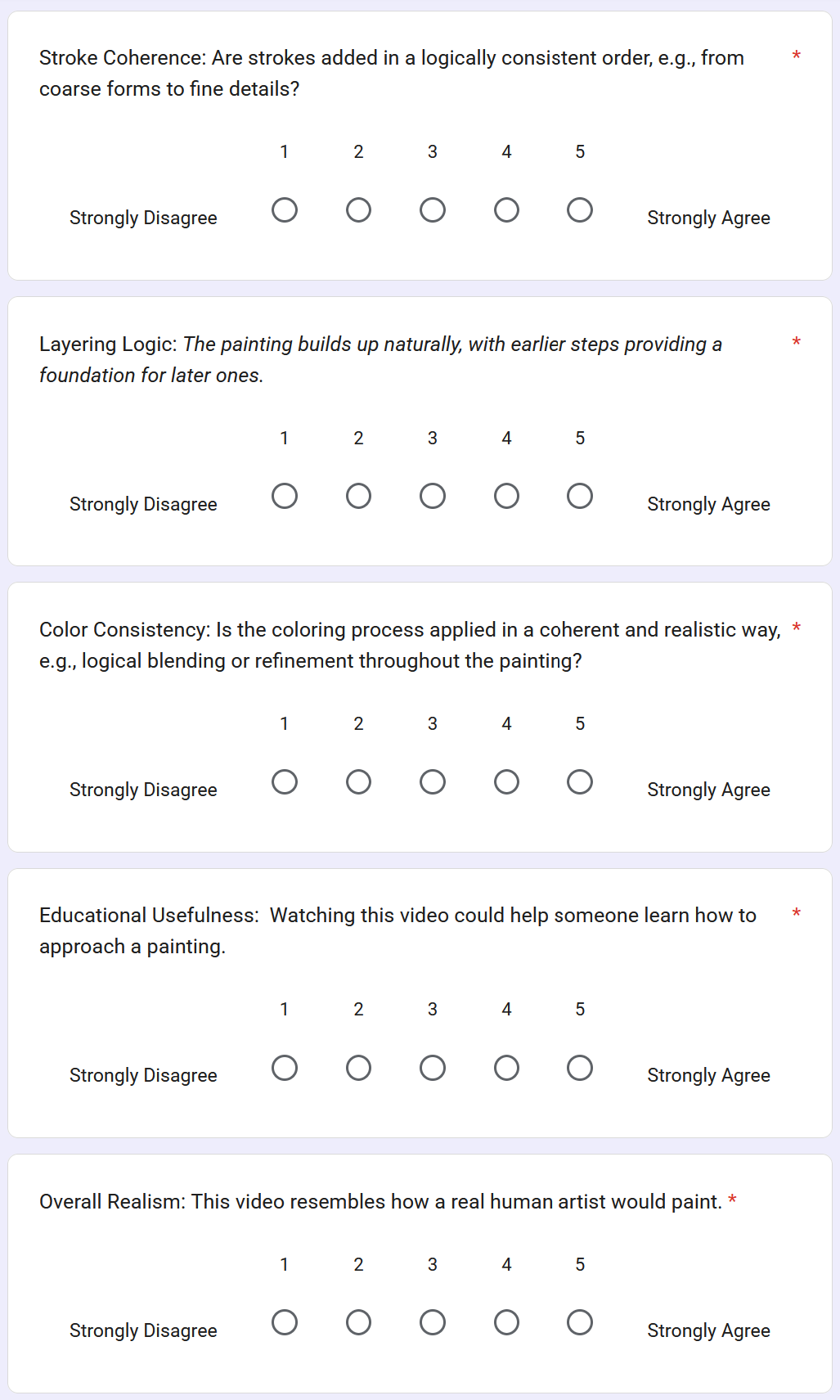}
    \vspace{-7mm}
    \caption{Screenshot of the expert user study. These questions were asked for each painting video.}
    \label{fig:appx:screenshot_expert_user}
    \vspace{-2mm}
\end{figure}

\subsection{Human evaluation}
\label{sec:appx:human_eval}

\subsubsection{Expert Evaluation Details.}
Expert participants were primarily recruited from specialized, professional artist communities on Reddit (e.g., r/artcommissions, r/oilpainting). To ensure high-quality feedback, we strictly selected artists who professionally commission or sell their artwork. 

\noindent As shown in \cref{fig:appx:screenshot_expert_user}, participants were asked to rate the following five criteria on a 5-point Likert scale:
\begin{itemize}
    \item \textbf{Stroke Coherence:} Are strokes added in a logically consistent order, e.g., from coarse forms to fine details?
    \item \textbf{Layering Logic:} The painting builds up naturally, with earlier steps providing a foundation for later ones.
    \item \textbf{Color Consistency:} Is the coloring process applied in a coherent and realistic way, e.g., logical blending or refinement throughout the painting?
    \item \textbf{Educational Usefulness:} Watching this video could help someone learn how to approach a painting.
    \item \textbf{Overall Realism:} This video resembles how a real human artist would paint.
\end{itemize}

\noindent The evaluation videos in \cref{tab:expert_study} (Videos 1--6) were sourced from \cite{aluart_stream}, \cite{onepencil2024girl}, \cite{samir2024lavender}, \cite{paintlane2018boats}, \cite{onepencil2024portrait}, and \cite{samir2024pine}, respectively.

\subsubsection{Human Preference Study}
In the user study, 22 participants evaluated pairwise video outputs across 25 test images based on two criteria: (1) accurately capturing the complete painting process, and (2) matching the reference image in the final frame. Participants could also declare a tie if both outputs were of comparable quality.

\begin{figure*}[htbp!]
    \centering
    \includegraphics[width=\textwidth]{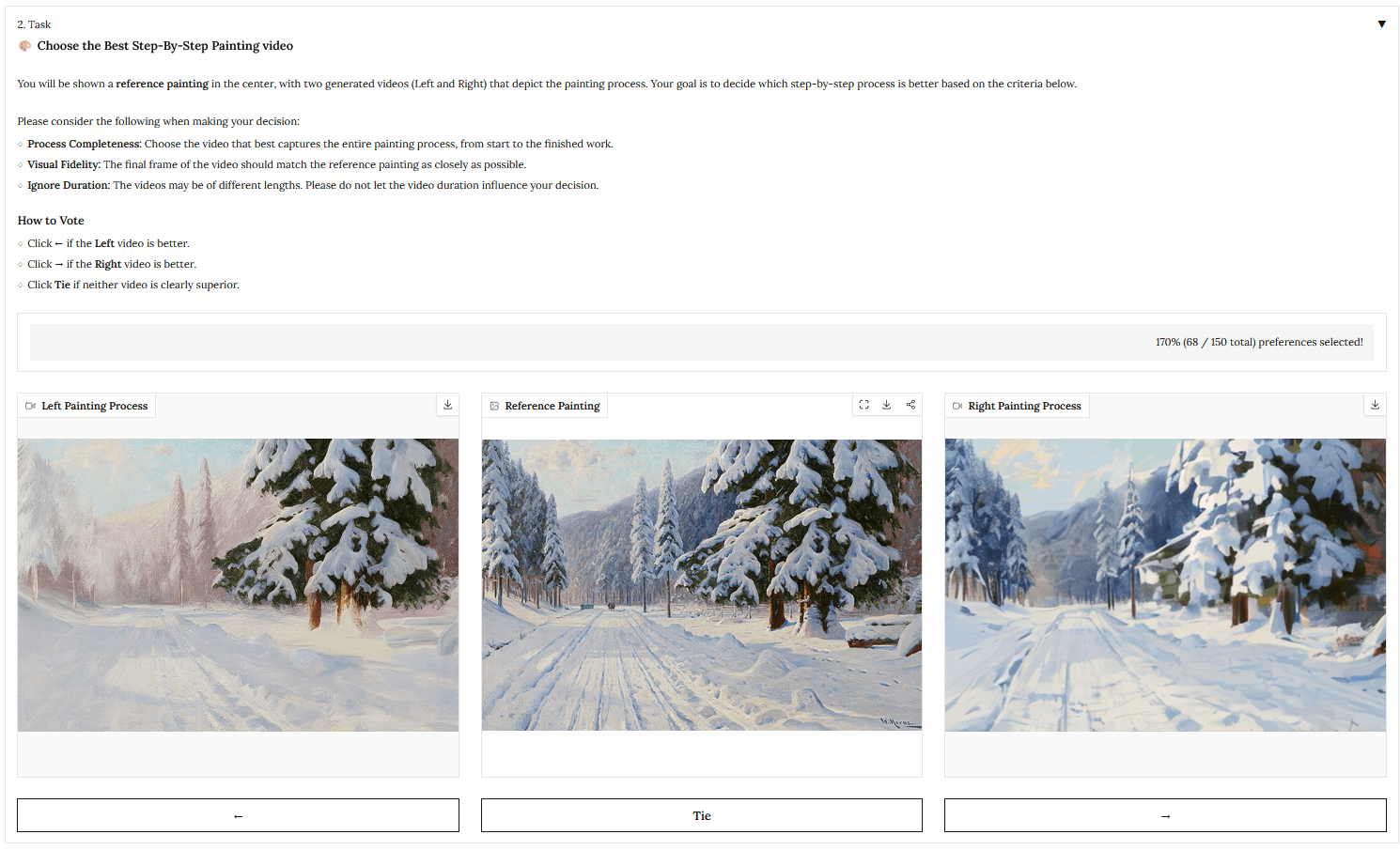}
    \caption{Screenshot of our human preference study.}
    \label{fig:appx:user_study_app}
\end{figure*}

Detailed results are reported in \cref{tab:appx:user_study}. Our method consistently outperforms all competing approaches ($p < 0.001$), securing win rates between 78.7\% and 96.2\% in pairwise comparisons. To analyze performance variations, \cref{tab:win_rates} breaks down win rates across specific top-performing and challenging images. While our method achieved unanimous preference (100\% win rate) on multiple samples, it maintained at least a 65\% win rate across all test cases. A screenshot of the evaluation interface is shown in \cref{fig:appx:user_study_app}.

\subsection{Quantitative Results}
Table \ref{tab:art_evaluation} provides a comparative analysis of our approach against state-of-the-art methods on the full test set. To account for potential performance biases stemming from the diversity of our data, we present a supplementary evaluation in \ref{tab:appx:art_evaluation} focusing exclusively on acrylic and oil paintings. The results demonstrate that the performance hierarchy remains consistent, with our method maintaining its lead across these specific media.

\begin{table}[htbp!]
    \centering
    \footnotesize
    \label{tab:efficiency}
    \begin{tabular}{lcccc}
        \toprule
        \textbf{Model} & \textbf{Params} & \textbf{Peak VRAM (GB)} & \textbf{Latency (s)} & \textbf{FPS} \\
        \midrule
        Wan 2.1 & 14B  & 47.0          & 428.3         & 0.19          \\
        Wan 2.2 & 5B   & 35.3          & \textbf{37.3} & \textbf{2.17} \\
        Vace    & 1.3B & \textbf{18.4} & 171.7         & 0.47          \\
        \bottomrule
    \end{tabular}
    \caption{\textbf{Inference Efficiency Comparison.} Benchmarked on a single NVIDIA A100 GPU generating 81 frames at $832\times480$ spatial resolution in \texttt{bf16} precision. Best results per column are highlighted in \textbf{bold}.}
\end{table}

\subsection{Inference Efficiency}
To evaluate the practical usability of our backbones, we benchmark their computational efficiency on a single NVIDIA A100 GPU (64GB). All models generate $81$ frames at $832 \times 480$ resolution using bfloat16 (\texttt{bf16}) precision with equivalent sampling step settings. We did not use any classifier free guidance since the prompt is irrelevant in this case, making the generation faster. As summarized in \cref{tab:efficiency}, Wan 2.2 (5B) achieves superior throughput, completing the 81-frame generation sequence in \textbf{37.3 seconds} ($2.17$ FPS). This represents an $11.5\times$ speedup over Wan 2.1 (14B) and a $4.6\times$ speedup over Vace (1.3B). While Vace yields the lowest peak VRAM footprint at $18.4$~GB (making it suitable for lower-tier consumer GPUs with 24GB VRAM), its architecture exhibits higher sampling latency per frame.
Wan 2.2 strikes the most effective trade-off between hardware accessibility ($35.3$~GB peak VRAM) and generation speed, making it the most suitable architecture for our pipeline. Its latency advantage stems from a more compressed latent space and an efficient conditioning mechanism: Wan 2.2 prepends the reference image directly to the token sequence. In contrast, the other backbones concatenate the reference image channel-wise by padding an entirely new video-sized latent tensor, which drastically increases the total latent size and computational complexity.

\begin{table*}[htpb!]
  \centering
  \begin{tabular}{ll ccc cc}
    \toprule
    \textbf{Method A} & \textbf{Method B} & \textbf{Win (A) (\%)} & \textbf{Tie (\%)} & \textbf{Win (B) (\%)} & \textbf{Total} & \textbf{$p$-value} \\
    \midrule
    Ours & PaintsUndo & 95.6 & 1.3 & 3.2 & 158 & $<0.001$ \\
    Ours & Inverse Painting & 78.7 & 9.8 & 11.6 & 164 & $<0.001$ \\
    Ours & ProcessPainter & 96.2 & 0.0 & 3.8 & 160 & $<0.001$ \\
    \midrule
    PaintsUndo & Inverse Painting & 27.2 & 14.2 & 58.6 & 162 & $<0.001$ \\
    PaintsUndo & ProcessPainter & 47.0 & 14.6 & 38.4 & 164 & 0.272 \\
    Inverse Painting & ProcessPainter & 63.4 & 8.5 & 28.0 & 164 & $<0.001$ \\
    \bottomrule
  \end{tabular}
  \caption{Human evaluation results for overall preference. \textbf{Method A} and \textbf{Method B} denote the two compared approaches. The $p$-values indicate statistical significance.}
  \label{tab:appx:user_study}
\end{table*}

\begin{table*}[t]
    \centering
    \begin{tabular}{l c c c c l c c c}
        \toprule
        \multicolumn{4}{c}{\textbf{Top Performing Images}} & & \multicolumn{4}{c}{\textbf{Most Challenging Images}} \\
        \cmidrule{1-4} \cmidrule{6-9}
        \textbf{Image} & \textbf{Matches} & \textbf{Wins} & \textbf{Win Rate} & & \textbf{Image} & \textbf{Matches} & \textbf{Wins} & \textbf{Win Rate} \\
        \midrule
        \cref{fig:art_media_comparison_000} (b) & 23 & 23 & 1.00 & & \cref{fig:appx:comparison_2} (c) & 20 & 13 & 0.65 \\
        \cref{fig:showcase} (a) & 22 & 22 & 1.00 & & \cref{fig:appx:comparison_2} (b) & 16 & 12 & 0.75 \\
        \cref{fig:appx:ours_1} (b) & 21 & 21 & 1.00 & & Not shown & 19 & 15 & 0.79 \\
        \cref{fig:appx:comparison_2} (a) & 20 & 20 & 1.00 & & \cref{fig:art_media_comparison_000} (a) & 20 & 16 & 0.80 \\
        \bottomrule
    \end{tabular}
    \caption{Overall user preference win rates for our model across all baselines. The left side highlights our top performing cases where human preference was unanimously in favor of our method (100\% win rate). The right side shows our most challenging edge cases, where despite struggling, our method maintained a robust minimum win rate of 65\%.}
    \label{tab:win_rates}
\end{table*}

\section{Dataset Curation Pipeline}
\label{appx:dataset_curation_pipeline}

A more in depth discussion about the dataset curation pipeline than in the main paper.

\subsection{Painting Video Extraction Pipeline}
\noindent\textbf{Temporal Trimming.}  Raw tutorial videos often include irrelevant introductory or outro segments. To isolate the painting process, we detect the start frame as the first occurrence of a hand (indicating artist activity) and the end frame as the last hand appearance using GroundingDINO \cite{liu2024grounding}. The video is then trimmed to this interval.

\noindent\textbf{Canvas Localization.} We localize the canvas using Grounding DINO’s zero-shot object detection, querying for “canvas”. For Loomis portrait tutorials, which typically display a reference photograph on the left and the canvas on the right, we compute horizontal intensity gradients across each frame and split the image at the column with maximal gradient magnitude, isolating the canvas region. This reliably isolates the canvas region, as the reference photograph typically has a dark monotone background, while the canvas is a white sheet, producing a strong gradient.

\noindent\textbf{Frame Sampling and Occlusion Removal}.
The trimmed video is partitioned into $N=\frac{video duration (sec)}{10}$ segments. From each segment, we sample $M=30$ frames (3 frames/sec). For each sampled frame we detect occlusions (e.g., hands, brushes) using InSPyReNet \cite{kim2022revisiting} or BiRefNet \cite{BiRefNet}, which segment foreground objects via iterative refinement. Afterwards a median frame is computed from the $M$ samples, masked to exclude occluded pixels. The mask ensures that none of the detected occlusions are part of the final frame. To fill regions persistently occluded in the sample, we iteratively include the median of prior segments in the computation, initializing with a blank white canvas. The process leaves us with $N$ frames.

\noindent\textbf{Post-Processing.} Logos and text overlays are detected using Grounding DINO and removed via inpainting with LaMa \cite{suvorov2022resolution}.

\noindent\textbf{Efficiency}. On an NVIDIA RTX A4000 GPU, our pipeline processes videos with a resolution of 640 $\times$ 360 pixels in near real-time (processing time approximately equals video duration), enabling scalable dataset curation.

\subsection{Data Collecting}
We curate a dataset from painting tutorial videos on YouTube, prioritizing videos with static camera angles and minimal canvas movement to simplify temporal alignment. After processing the videos using the pipeline introduced in Sec.~\ref{appx:dataset_curation_pipeline}, we manually reviewed the outputs and excluded those with poor results. In total, we collected 737 videos spanning a variety of art media and artists.

\noindent\textbf{Acrylic.} This subset includes 81 photorealistic acrylic landscape painting tutorials (avg. 40 min duration), emphasizing techniques like wet-on-wet blending and layering.

\noindent\textbf{Oil.}  We collected 151 oil painting tutorials, including 142 impressionist landscapes (loose brushwork, vibrant palettes) and 9 photorealistic paintings (avg. 30 min duration).

\noindent\textbf{Pencil.} This subset comprises 270 pencil and 28 colored pencil tutorials (avg. 12 min duration), covering various scenes and motives.

\noindent\textbf{Loomis Portraits.} We include 207 portrait tutorials following Andrew Loomis’ proportional method \cite{loomis2021drawing}. These videos typically display a reference photograph on the left and the canvas on the right. We isolate the canvas via horizontal gradient splitting, as described in Sec. \ref{appx:dataset_curation_pipeline}.


\subsection{Painting Media Transfer Dataset}
\label{appx:painting_media_transfer_dataset}
Unlike the combined dataset, which uses finished paintings as references, Loomis tutorials require conditioning on real portrait photographs rather than sketches. Similar we want to enable media transfer for the other medias. Given a sketch, generate a oil or acrylic painting and vice versa. To bridge this gap, we synthesize variations fo the reference frame with image editing models. For the pencil sketches and loomis portrait drawings we generate realistic photos and color paintings using ControlNet, ensuring alignment between reference photos and artwork. For acrylic and oil paintings we use Qwen Image Edit \cite{wu2025qwen} to generate pencil sketches, children drawings and drawing book styles.

\noindent The tradeoff we noticed is that ControlNet follows exactly the outlines of the reference paintings but looses information such as color due to the underlying control mechanism. On the other side Image Editing models keep this information but the ones we tested slightly change the image composition either due to the models generative capabilities or due to a resolution mismatch. Therefore we used ControlNet for pencil sketches and Qwen Image Edit for the color paintings.

\noindent When we generate reference frame variations with ControlNet we use the LineArt processor, it works well for paintings. \cite{epicrealism2023} was used as the image generation model, it is a fine tuned version of Stable Diffusion 1.5 \cite{rombach2022high}.

\noindent We generate captions for the original reference frame with LLavaNexT \cite{liu2023improved}, these captions get combined with the art media label. During fine tuning we replace the original frame with the generated frames and the model has to learn the transfer based on the prompt.

\begin{table*}[htpb!]
    \centering
    \small
    \begin{tabular}{l|ccccc|ccccc}
        \toprule
        & \multicolumn{5}{|c}{Acrylic} & \multicolumn{5}{|c}{Oil} \\
        Method & \multicolumn{1}{|c}{FID $\downarrow$} & \multicolumn{1}{c}{LPIPS $\downarrow$} & \multicolumn{1}{c}{Clip $\uparrow$} & \multicolumn{1}{c}{Dinov2 $\uparrow$} & \multicolumn{1}{c}{DDC $\downarrow$} & \multicolumn{1}{|c}{FID $\downarrow$} & \multicolumn{1}{c}{LPIPS $\downarrow$} & \multicolumn{1}{c}{Clip $\uparrow$} & \multicolumn{1}{c}{Dinov2 $\uparrow$} & \multicolumn{1}{c}{DDC $\downarrow$} \\
        \midrule
        Inverse Painting ~\cite{chen2024inverse} & 287.58 & 0.77 & 0.69 & 0.27 & 88.76 & 292.05 & 0.77 & 0.67 & 0.27 & 69.39 \\
        ProcessPainter~\cite{song2024processpainter} & 223.16 & 0.60 & 0.77 & 0.51 & 38.44 & 257.34 & 0.60 & 0.75 & 0.55 & 21.89 \\
        PaintsUndo~\cite{paintsundo} & 209.22 & 0.66 & 0.76 & 0.50 & 36.10 & 213.69 & 0.60 & 0.77 & 0.58 & 23.15 \\
        PaintsUndo$^\dagger$~\cite{paintsundo} & \underline{152.90} & \underline{0.50} & \underline{0.83} & \underline{0.69} & \underline{16.91} & \underline{170.84} & \underline{0.47} & \underline{0.86} & \underline{0.75} & \underline{13.73} \\
        \midrule
        Ours & \textbf{104.37} & \textbf{0.44} & \textbf{0.89} & \textbf{0.81} & \textbf{16.18} & \textbf{145.02} & \textbf{0.45} & \textbf{0.88} & \textbf{0.79} & \textbf{8.47} \\
        \bottomrule
    \end{tabular}
    \vspace{-2mm}
    \caption{Evaluation of the Acrylic and Oil painting subsets from our test set. Methods marked with $^\dagger$ were fine-tuned on our proposed dataset. \textbf{Bold} and \underline{underline} indicate the best and second-best results, respectively. Overall, our method consistently outperforms both zero-shot and fine-tuned baselines across metrics and art media.}
    \label{tab:appx:art_evaluation}
    \vspace{10mm}
\end{table*}

\end{document}